\documentclass[review,12pt]{elsarticle}

\usepackage{amssymb}
\usepackage{amsmath}
\usepackage{subfigure}
\usepackage{algorithm}
\usepackage{algpseudocode}
\usepackage{xcolor}
\usepackage{tikz}
\usepackage{subcaption}
\usepackage{graphicx}
\usepackage{pdflscape}

\newcommand{\hdtwo}[2]{{\renewcommand{\arraystretch}{0.8}\begin{tabular}[b]{@{}c@{}}#1\\#2\end{tabular}}}
\newcommand{\lbtwo}[2]{{\renewcommand{\arraystretch}{0.8}\begin{tabular}[t]{@{}l@{}}#1\\#2\end{tabular}}}
\usetikzlibrary{arrows.meta,positioning,fit,backgrounds,calc,shapes.geometric}

\usepackage{lineno}
\journal{Engineering Applications of Artificial Intelligence}
\begin{document}

\begin{frontmatter}



\title{Hierarchical Interpolating MeshGraphNets: Accurate and Efficient Mesh-based Physics Learning with Hierarchical Multi-scale Graph Neural Networks} 


\author[1]{SiHun Lee}
\ead{s.hun.lee@samsung.com}
\author[2]{Dong-Hyuk Park}
\author[2]{Taesoo Bang}
\author[2]{Seung-Hoon Kang\corref{cor1}}
\ead{shkang@sejong.ac.kr}
\cortext[cor1]{Corresponding author}

\affiliation[1]{organization={Mobile Experience, Digital Twin AX Lab., Samsung Electronics Co.},
            addressline={129, Samsung-ro, Yeongtong-gu}, 
            city={Suwon-si},
            postcode={16677},
            state={Gyeonggi-do},
            country={Republic of Korea}}  
\affiliation[2]{organization={Department of Aerospace System Engineering, Sejong University},
            addressline={209, Neungdong-ro, Gwangjin-gu}, 
            city={Seoul},
            postcode={05006},
            country={Republic of Korea}}

\begin{abstract}
Machine-learned physics surrogate models have become promising alternatives to mesh-based numerical solvers. Among them, graph neural networks (GNNs) have been successful in representing simulation meshes as graphs and learning nodal state evolution via message passing. However, conventional message passing can become inefficient for large, high-fidelity meshes. Since information propagates only one hop per step, long-range interactions require many steps, increasing computational burden, memory usage, and the risk of over-smoothing.

We propose Hierarchical Interpolating MeshGraphNets (HI-MGN), a new multiscale graph neural network that improves long-range communication on large meshes.
HI-MGN replaces the flat processor with a hierarchical multiscale processor. On each level, farthest point sampling and graph-based Voronoi partitioning divide the mesh into a user-set number of clusters. Each cluster becomes a coarse node, and two clusters are connected only if an original mesh edge joins them, so no artificial connections are introduced. These coarse graphs enable information to propagate over larger geometric distances with fewer message passing steps. The coarse states are projected back onto the original mesh via a learned interpolation GNN, where every fine node receives messages from its own cluster and the adjacent clusters.

On structural and fluid problems, HI-MGN achieves the best overall accuracy against four GNN baselines while remaining computationally efficient. On the full NASA Common Research Model dataset, HI-MGN yields a lower mean error than those reported for existing methods. These results suggest that topology-aware hierarchical processing and learned coarse-to-fine interpolation provide an effective and practical route toward scalable mesh-based physics surrogate modeling.
\end{abstract}


\begin{highlights}
\item HI-MGN implements FPS-Voronoi multiscale techniques for mesh-based GNNs.
\item FPS-Voronoi coarsening follows mesh adjacency; learned interpolation restores detail.
\item Best overall accuracy against four GNN baselines on structural and fluid problems.
\item HI-MGN remains computationally efficient in GPU memory and training time.

\end{highlights}


\end{frontmatter}



\section{Introduction}\label{sec:1}

With the rapid progress of artificial intelligence (AI) and machine learning, surrogate models that learn to predict the results of physics simulations have attracted growing attention. This research area is known as scientific machine learning (SciML) or AI for computer-aided engineering (AI-CAE). Such surrogate models yield predictions close to the results of full numerical simulations at a fraction of their computational cost. Early surrogates mainly applied classical regressors, such as Gaussian process and radial basis function regression, to linear reduced bases from proper orthogonal decomposition \cite{ostrowski2008solving,guo2018reduced}. Later, the regressors on such bases were replaced by neural networks, such as feed-forward networks \cite{Hesthaven:2018a,li2021nonintrusive,kneifl2021nonintrusive}, long short-term memory (LSTM) networks \cite{mohan2018deep}, variational autoencoders (VAEs) \cite{Lee:2024a}, and generative adversarial networks (GANs) \cite{Lee:2021a}. In addition, autoencoders \cite{Kadeethum:2022a,Kadeethum:2022b,xu2020multi,lee2020model} and VAEs \cite{solera2023beta,kang2022physics,wang2021flow,Lee:2024b,lee2025physics} replaced the linear basis itself with a nonlinear encoder and decoder. These models work well on structured grids or a fixed discretization, but their inputs and outputs are tied to one set of degrees of freedom. Therefore, they are difficult to apply to unstructured meshes or changing geometries and were mainly applied to differing boundary conditions.

Mesh-agnostic models, which can be trained and evaluated on different meshes and geometry, remove this restriction. Early examples of such models include PointNet \cite{qi2017pointnet} and PointNet++ \cite{qi2017pointnet++}. PointNet applies a shared network to each point and combines the results by symmetric pooling, while PointNet++ applies it hierarchically to local neighborhoods. However, these models ignore mesh connectivity and can lose local detail on large or unevenly sampled meshes. Neural operators, another mesh-agnostic approach, learn maps between functions. DeepONet \cite{lu2021learning} uses separate networks for the input function and the query coordinates, and Geo-FNO \cite{li2023fourier} deforms an irregular domain onto a uniform grid. The geometry-informed neural operator (GINO) \cite{li2023geometry} combines graph and Fourier layers for large 3D problems. However, these operator-learning models can struggle with large variation in geometries and boundary conditions, fine geometric features, sharp local gradients or shocks, and extrapolation.

Transformer-based solvers such as Transolver \cite{wu2024transolver}, Transolver++ \cite{luo2025transolver++}, Transolver-3 \cite{zhou2026transolver} and AB-UPT \cite{alkin2025ab} exchange nodal information by attention instead of along mesh edges. They are currently among the most accurate surrogates on industrial aerodynamic benchmarks, but on large meshes they can still require a extensive amount of memory and computation.

A simulation mesh is naturally a graph, which motivates the use of graph neural networks (GNNs) on meshes. One of the first successful GNNs for numerical simulation is MeshGraphNets (MGN) \cite{pfaff2020learning}. It uses an encoder--processor--decoder structure, in which the processor passes messages between neighboring nodes. It showed that GNNs can learn how physical fields evolve on unstructured meshes, and many later models build on this structure. However, its processor is flat: it moves information by only one mesh edge per message passing step, so two nodes $r$ edges apart need at least $r$ steps to interact. Stacking many steps requires a large amount of memory and tends to over-smooth the nodal states. A deep flat GNN for distant data can therefore lose accuracy on contrary \cite{fortunato2022multiscale}.

Hierarchical GNNs address this limitation by also passing messages on coarser graphs, where a few steps carry long-range information across the domain. The coarse graphs may be supplied as auxiliary inputs or computed by coarsening the mesh itself. For example, MultiScale MGN \cite{fortunato2022multiscale} and GraphCast \cite{lam2023learning} use coarser meshes that are generated in advance, separately from the simulation mesh, and must be supplied for both training and inference. Other models build the hierarchy by coarsening the simulation mesh itself, so that each coarse node is either a selected mesh node or a cluster of mesh nodes. Models of this type include MultiScaleGNN \cite{lino2021simulating}, its successor REMuS-GNN \cite{lino2022multi}, the model of Han et al.\ \cite{han2022predicting}, the Bi-Stride Multi-Scale GNN (BSMS-GNN) \cite{cao2023efficient}, the hierarchical contact mesh transformer (HCMT) \cite{yu2024learning}, MAgNET \cite{deshpande2024magnet}, EvoMesh \cite{deng2025evomesh}, and physics-informed coarsening \cite{bazzi2026physics}. Recently, the diffusion graph network (DGN) \cite{lino2025learning} and ROBIN \cite{wurth2025diffusion} have combined such hierarchies with denoising diffusion.

Most of these methods use coarsening rules to build hierarchies. BSMS-GNN uses bi-stride pooling, roughly halving the node count on every level except the last \cite{cao2023efficient}, and DGN uses Guillard's coarsening, which reduces it about fourfold on 2D meshes \cite{lino2025learning}. ROBIN uses the root-node aggregation of algebraic multigrid, where the user chooses only the maximum size of the coarsest level, and the number of levels and the level sizes follow from the aggregation \cite{wurth2025diffusion}. With a fixed coarsening ratio, the user cannot freely choose the number of nodes on each level. The resulting hierarchy may therefore be poorly suited to certain meshes. In particular, a large mesh reaches a small coarsest level only through many levels, and a deeper hierarchy is not necessarily more accurate. This is because each added level costs message passing steps and transfers between levels, which are prone to downsampling and upsampling errors. It has also been reported that the best hierarchy depth depends on the mesh \cite{yu2024learning,deng2025evomesh,gao2019graph}.

Instead, the hierarchy may be set by the user. Several existing methods allow such user-set hierarchies including MultiScaleGNN that merges nodes that fall in the same cell of a regular grid with a user-set cell size \cite{lino2021simulating}. Han et al.\ keep a uniform sample of the nodes \cite{han2022predicting}, while Graph U-Nets keep the nodes with the largest learned scores \cite{gao2019graph}. Physics-informed coarsening keeps the nodes with the largest residuals of the governing equations and may connect them to their nearest neighbors \cite{bazzi2026physics}. However, grid cells and proximity-based connections, such as nearest-neighbor connections or the Delaunay connections of HCMT \cite{yu2024learning}, do not preserve mesh topology. They can merge or connect nodes on adjacent bodies or across a slot or crack, allowing information to travel along paths that do not exist in the mesh.

Existing methods also differ in how coarse states are upsampled back to the mesh. MAgNET and Graph U-Nets copy each coarse state to one fine node or to every node of its cluster \cite{deshpande2024magnet,gao2019graph}. MultiScaleGNN and DGN send a learned message from a single coarse node \cite{lino2021simulating,lino2025learning}, while Han et al.\ and BSMS-GNN interpolate from several coarse nodes with fixed weights \cite{han2022predicting,cao2023efficient}. MultiScale MGN, GraphCast, EvoMesh, and ROBIN use learned upsampling from several coarse nodes \cite{fortunato2022multiscale,lam2023learning,deng2025evomesh,wurth2025diffusion}. It has been reported that such learned upsampling reduces rollout error compared to fixed weights \cite{cao2023efficient,deng2025evomesh}.

Table~\ref{tab:hier_props} summarizes these differences as five properties. (i) \textit{No external coarse graph}: the coarse levels are computed from the original mesh, not supplied as a separate graph. (ii) \textit{Configurable hierarchy}: the level sizes are set by the user in the form of node counts, reduction ratios, or cell sizes, not by the coarsening rule. (iii) \textit{Topology-preserving coarsening}: fine nodes are merged and coarse nodes are connected only along paths of mesh edges, so that no artificial connections are generated between adjacent bodies or across a gap. (iv) \textit{Learned multi-source upsampling}: the state of each fine node is reconstructed from learned messages sent by several coarse nodes. (v) \textit{Single-pass inference}: each prediction takes one forward pass, without an iterative process such as denoising diffusion.

In this work, we propose \textit{Hierarchical Interpolating MeshGraphNets (HI-MGN)}. HI-MGN preserves the encoder--processor--decoder structure of MGN, but replaces the flat processor with a hierarchical one. The hierarchical processor performs message passing on the original mesh and its coarse graphs, which are built by topology-preserving coarsening. The state of each fine node is then reconstructed from several coarse nodes by a learned interpolation. In Table~\ref{tab:hier_props}, HI-MGN is compared with representative hierarchical GNNs in terms of the five properties defined above.

\begin{table}[htbp]
    \centering
    \footnotesize\setlength{\tabcolsep}{1.6pt}\begin{tabular}{@{}l|ccccc@{}}
        \hline
        Method & \shortstack{No external\\coarse graph} & \shortstack{Configurable\\hierarchy} & \shortstack{Topology-\\preserving\\coarsening} & \shortstack{Learned\\multi-source\\upsampling} & \shortstack{Single-\\pass\\inference} \\
        \hline
        Graph U-Nets (2019) \cite{gao2019graph} & \checkmark & \checkmark & \checkmark & $\times$ & \checkmark \\
        MultiScaleGNN (2021) \cite{lino2021simulating} & \checkmark & \checkmark & $\times$ & $\times$ & \checkmark \\
        Han et al.\ (2022) \cite{han2022predicting} & \checkmark & \checkmark & $\times$ & $\times$ & \checkmark \\
        MultiScale MGN (2022) \cite{fortunato2022multiscale} & $\times$ & \checkmark & $\times$ & \checkmark & \checkmark \\
        BSMS-GNN (2023) \cite{cao2023efficient} & \checkmark & $\times$ & \checkmark & $\times$ & \checkmark \\
        GraphCast (2023) \cite{lam2023learning} & $\times$ & \checkmark & $\times$ & \checkmark & \checkmark \\
        MAgNET (2024) \cite{deshpande2024magnet} & \checkmark & $\times$ & \checkmark & $\times$ & \checkmark \\
        DGN (2025) \cite{lino2025learning} & \checkmark & $\times$ & \checkmark & $\times$ & $\times$ \\
        ROBIN (2025) \cite{wurth2025diffusion} & \checkmark & $\times$ & \checkmark & \checkmark & $\times$ \\
        HI-MGN (ours) & \checkmark & \checkmark & \checkmark & \checkmark & \checkmark \\
        \hline
    \end{tabular}
    \caption{Comparison of hierarchical graph neural networks.}
    \label{tab:hier_props}
\end{table}

We evaluate HI-MGN against four GNN baselines, MGN \cite{pfaff2020learning}, BSMS-GNN \cite{cao2023efficient}, ROBIN \cite{wurth2025diffusion}, and the Vanilla GNN of DGN \cite{lino2025learning}, on two numerical examples: 3D nonlinear structural contact and 3D aerodynamics of the NASA-CRM benchmark. The contributions of this work are as follows:
\begin{itemize}
    \item \textit{User-set hierarchy with topology-preserving coarsening.} The number of nodes on every level is set by the user, not by the coarsening rule, and the coarse graphs are computed from the mesh itself. Two coarse nodes are connected only where a mesh edge joins their clusters, so no connections are introduced between separate bodies. The ablation study (Section~\ref{sec:ablation}) confirms that removing this hierarchy substantially reduces accuracy, and that the accuracy also depends on the number of hierarcy levels.
    \item \textit{Learned GNN interpolation from adjacent clusters.} Each fine node is reconstructed by learned messages from its own cluster and from every cluster adjacent to it in the coarse graph. In the ablation study (Section~\ref{sec:ablation}), this yields a higher mean $R^2$ than a linear interpolator and a learned interpolation from the node's own cluster alone.
    \item \textit{Highest accuracy among the compared models while remaining computationally efficient.} As a result, HI-MGN yields the highest mean $R^2$ among the four baselines on both the structural contact and aerodynamic problems, while still remaining efficient in terms of peak GPU memory (VRAM) usage and training time. In particular, on the full aerodynamic dataset of the NASA Common Research Model (NASA-CRM), HI-MGN trained on a single consumer-grade GPU yields a lower mean error than those reported for existing methods, including recent transformer-based models such as Transolver-3 and AB-UPT (Section~\ref{subsec:crm_full}).
\end{itemize}

MeshGraphNets is implemented in the same code base as HI-MGN, while BSMS-GNN \cite{cao2023efficient}, the diffusion-based ROBIN \cite{wurth2025diffusion}, and the deterministic Vanilla GNN of the DGN code \cite{lino2025learning} are implemented using their authors' code. The remainder of this paper is organized as follows. Section~\ref{sec:method} describes the architecture, hierarchical processor, and training of HI-MGN. In Section~\ref{sec:examples}, two numerical examples are presented with a comparison of HI-MGN to the baselines. In Section~\ref{sec:ablation}, an ablation study on the contact benchmark examines the hierarchy, message passing blocks, pooling and interpolation rules, and positional features. Finally, conclusions are given in Section~\ref{sec:conclusions}.

\section{Methodology}\label{sec:method}

\subsection{Model definition}
\label{subsec1}

HI-MGN follows the basic structure of MeshGraphNets \cite{pfaff2020learning}. We model a temporally discretized numerical simulation as a sequence of graphs,
\begin{equation}
  \{G^{0},G^{1},\dots,G^{T}\},\qquad G^{t}=(V^{t},E^{t}),\qquad E^{t}=E^{M}\cup E^{W,t},
  \label{eq:graph_seq}
\end{equation}
where $G^{t}$ denotes the graph at time step $t$, $T$ is the final time step, and the edge set $E^{t}$ consists of two types of bidirectional edges, mesh edges $E^{M}$ and world edges $E^{W,t}$. The nodes of $G^{t}$ are the nodes of the simulation mesh, and $G^{t}$ is hereafter referred to as the input graph. Each node in the node set $V^{t}$ stores nodal information such as deformation, physical quantities of interest, and categorical tags including boundary conditions, part numbers, and material types.

HI-MGN predicts the change of the nodal physical state. Denoting the physical state of node $i$ at time $t$ by $\mathbf{u}_{i}^{t}$, the network predicts
\begin{equation}
  \Delta\mathbf{u}_{i}^{t}=
  \begin{cases}
    \mathbf{u}_{i}^{t+1}-\mathbf{u}_{i}^{t}, & \text{transient case},\\
    \mathbf{u}_{i}-\mathbf{u}_{i}^{0}, & \text{static case},
  \end{cases}
  \label{eq:target}
\end{equation}
where $\mathbf{u}_{i}$ is the static solution. In the static case, the network takes $G^{0}$ as input and predicts the difference of the nodal state from the reference state $\mathbf{u}_{i}^{0}$. In compact form, HI-MGN maps the graph to the set of nodal state changes as
\begin{equation}
  \mathcal{F}_{\theta}:\;G^{t}\;\rightarrow\;\{\Delta\mathbf{u}^{t}_{i}\}_{i\in V^{t}},
  \label{eq:map}
\end{equation}
where $\mathcal{F}_{\theta}$ denotes the network with trainable parameters $\theta$, which consists of the encoder, hierarchical processor, and decoder.

\subsection{Architecture}
\label{subsec:architecture}

HI-MGN uses the standard `Encoder--Processor--Decoder' structure adopted by previously developed GNN surrogate models \cite{battaglia2018relational,sanchez2020learning,pfaff2020learning}. However, HI-MGN replaces the conventional flat processor with a hierarchical multiscale processor. The encoder and decoder follow the same formulation as those of MeshGraphNets and are summarized below \cite{pfaff2020learning}. Throughout the model, multi-layer perceptrons (MLPs) with two hidden layers and SiLU activations are used, and LayerNorm is applied to their outputs except in the decoder. All latent states have the same dimension $d_z$, and square brackets denote the concatenation of vectors.

\paragraph{Encoder} A nodal encoder $\mathbb{E}_{V}$ maps each nodal feature, consisting of the normalized input channels, positional features, and one-hot encoded categorical tags, to a node latent state $z_i$,
\begin{equation}
  z_i = \mathbb{E}_{V}(v_i),
\end{equation}
where $v_i$ denotes the nodal feature of the $i$-th node. Input channels are nodal quantities given to the network, such as the current physical state $\mathbf{u}^{t}_{i}$ and problem conditions.

Positional features are computed for each node from reference positions and mesh edges $E^{M}$. For node $i$, the positional features are
\begin{equation}
  \Big[\,\lVert p_{i}-\bar{p}\rVert,\;\;
  \ell_{i},\;\;
  \mathrm{RW}_{ii}^{2},\;\;
  \mathrm{RW}_{ii}^{4}\,\Big],
  \label{eq:posfeat}
\end{equation}
where $p_{i}$ is the reference position of node $i$, $\bar{p}$ is the mean reference position of all nodes, and $\ell_{i}$ is the mean length of mesh edges connected to node $i$. These are the distance of node $i$ from the graph centroid and the local element size, respectively. The latter two features are the random-walk positional encoding (RWPE) of Dwivedi et al.~\cite{dwivedi2022graph}, where $\mathrm{RW}_{ii}^{m}$ is the probability that a random walk along the mesh edges returns to node $i$ after $m$ steps.

Mesh edges $E^{M}$ follow the connectivity of the simulation mesh. Each mesh edge stores a relative geometric vector and its Euclidean norm between the connected nodes $i$ and $j$, evaluated in both the configuration at time $t$ and the reference configuration. Thus, the mesh-edge features $e^{M}_{ij}$ are defined as
\begin{equation}
  e^{M}_{ij}=\big[\,p^{t}_{i}-p^{t}_{j},\;\lVert p^{t}_{i}-p^{t}_{j}\rVert,\;p_{i}-p_{j},\;\lVert p_{i}-p_{j}\rVert\,\big],
  \label{eq:edge_feat}
\end{equation}
where $p^{t}_{i}$ denotes the current position of node $i$. World edges $E^{W,t}$ are constructed dynamically at each time step, connecting nodes whose current separation in the Lagrangian description falls below a prescribed threshold. These edges are intended to represent non-mesh interactions such as contact in structural dynamics and are constructed only on the input graph. World-edge features $e^{W}_{ij}$ are defined in the same way as $e^{M}_{ij}$ in Eq.~\eqref{eq:edge_feat}.

Separate mesh ($\mathbb{E}_{M}$) and world edge encoders ($\mathbb{E}_{W}$) embed the edge features of the corresponding two edge types. The corresponding latent states become
\begin{equation}
  z^{M}_{ij} = \mathbb{E}_{M}\!\big(e^{M}_{ij}\big), \qquad
  z^{W}_{ij} = \mathbb{E}_{W}\!\big(e^{W}_{ij}\big).
\end{equation}

\paragraph{Message passing block} A message passing (MP) block first updates the edge latent states based on the latent states of their corresponding sender and receiver nodes. Then, the nodal latent states are updated using the aggregated incoming edge messages. The update procedures are expressed as
\begin{align}
  \mu^{M}_{ij} &{}=\phi_{M}\big([\,z_i,\,z_j,\,z^{M}_{ij}\,]\big),\qquad
    z^{M}_{ij}\leftarrow z^{M}_{ij}+\mu^{M}_{ij},
    \label{eq:edge_update}\\[2pt]
  z_i &{}\leftarrow z_i+\phi_{V}\Big(\big[\,z_i,\;\textstyle\sum_{j}\mu^{M}_{ij},\;
    \textstyle\sum_{j}\mu^{W}_{ij}\,\big]\Big),
    \label{eq:node_update}
\end{align}
where $\phi_M$ and $\phi_V$ are independently parameterized MLPs and $\mu^{M}_{ij}$ is the message of the mesh edge from sender node $j$ to receiver node $i$. World-edge latent states, if present, are updated in the same way by $\phi_{W}$ and enter the node update through the aggregated messages $\sum_{j}\mu^{W}_{ij}$. The two sums are taken over nodes $j$ connected to node $i$ by a mesh edge and by a world edge, respectively.

\paragraph{Decoder} The decoder network $\mathbb{D}$ maps the refined message-passed node latent state of each node to a normalized per-node physical state update,
\begin{equation}
  \hat{y}_i = \mathbb{D}(z_i),
\end{equation}
where $\hat{y}_i$ is the prediction of the state change of Eq.~\eqref{eq:target} in normalized form. The predicted update is then denormalized and accumulated onto the current physical state to obtain the next state,
\begin{equation}
  \mathbf{u}^{t+1}_i = \mathbf{u}^{t}_i + \mathcal{N}^{-1}(\hat{y}_i),
\end{equation}
where $\mathcal{N}$ denotes the normalization operator. At inference, this update is applied autoregressively, and the edge features and world edges are recomputed from the predicted state at each time step.

\subsection{Hierarchical multiscale processor}
\label{subsec:vcycle}

\begin{figure}[h]
\centering

\includegraphics[
    width=\linewidth,
    height=\textheight,
    keepaspectratio
]{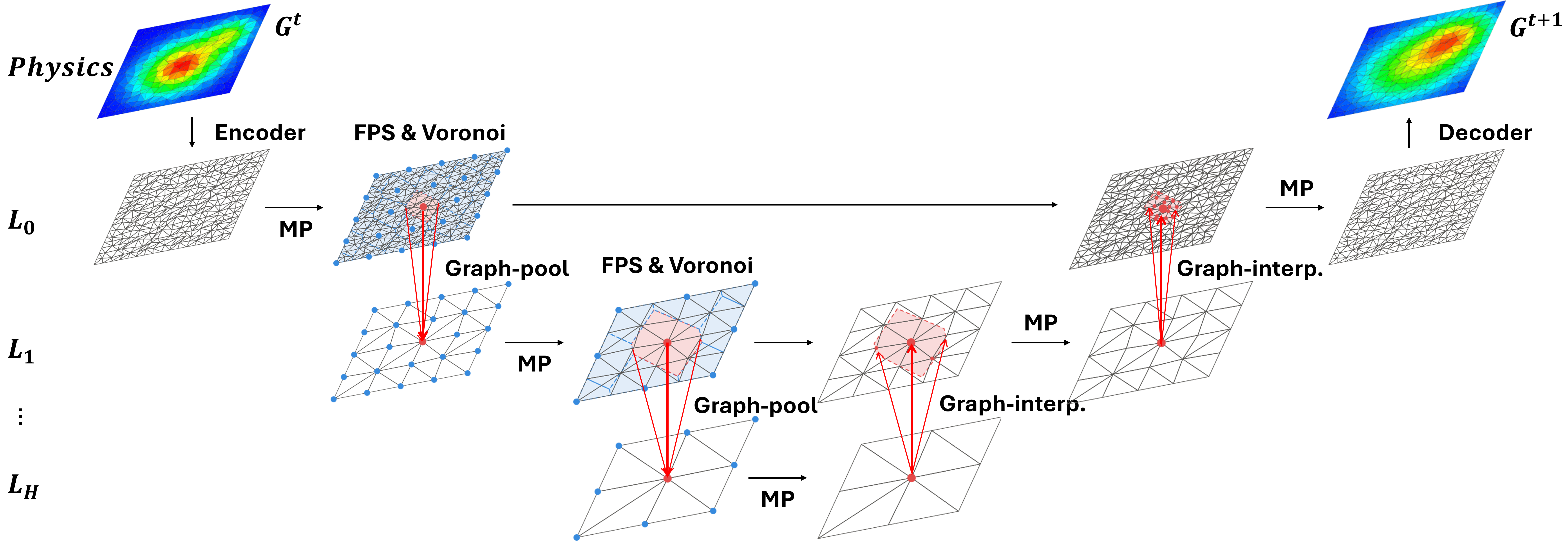}

\caption{Architecture of HI-MGN with hierarchical, multiscale message passing.}

\label{fig:architecture}
\end{figure}

Rather than stacking message passing blocks on the input graph, HI-MGN arranges message passing in a hierarchical manner over $H$ coarse graphs as shown in Fig.~\ref{fig:architecture}. In a flat processor, long-range interaction may require a prohibitively deep network that will frequently result in over-smoothing on large or high-fidelity meshes \cite{li2018deeper}. A multiscale processor shortens these communication paths by performing message passing on coarser graphs.

A hierarchical multiscale processor consisting of $H$ coarse levels is considered, where level $0$ is the input graph (original mesh) and levels $1,\dots,H$ are the coarse graphs. As illustrated in Fig.~\ref{fig:architecture}, the processor first applies message passing blocks on the input graph. The latent states are then pooled onto the next coarse graph and refined by message passing blocks. This process is repeated down to the coarsest level. On the way up, the latent states are interpolated onto the next finer level, merged with the stored skip states, and refined by message passing blocks again. The final latent states on the input graph are passed to the decoder. The numbers of message passing blocks are written as a list of $2H+1$ entries in the order of application.

\begin{figure}[h]
\centering

\includegraphics[
    width=\linewidth,
    height=\textheight,
    keepaspectratio
]{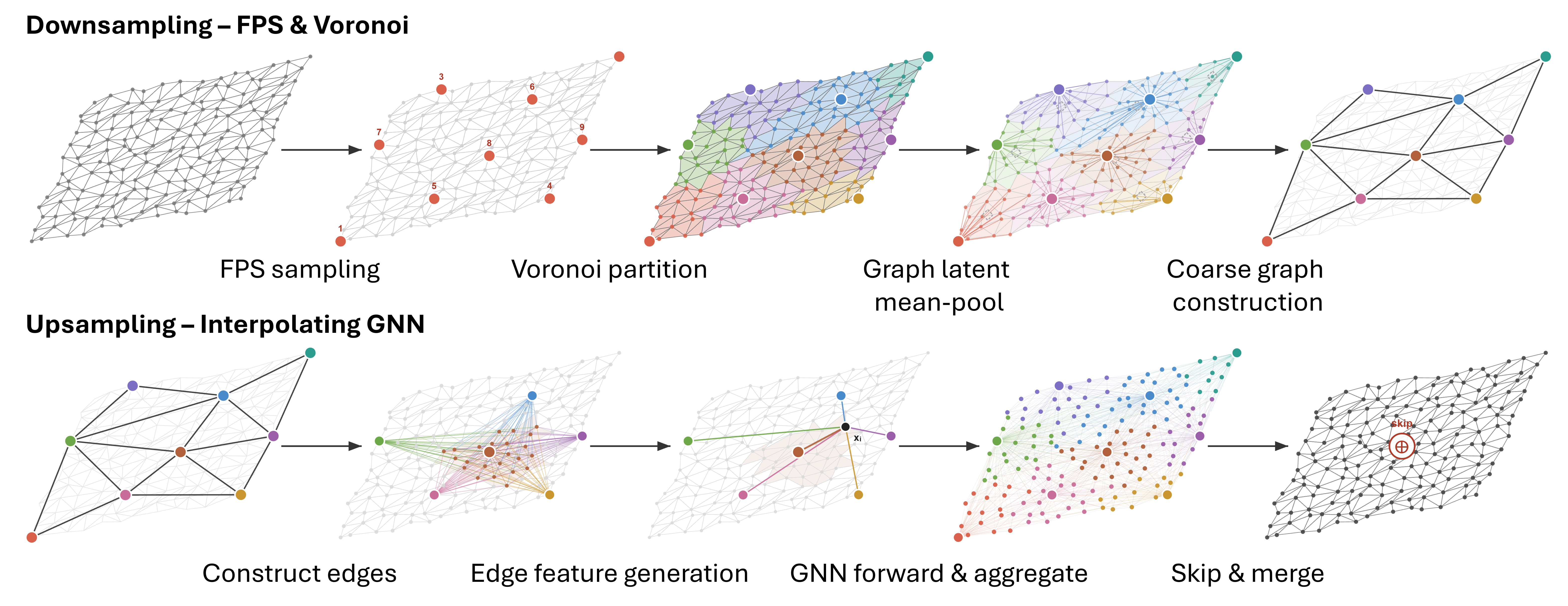}

\caption{Coarsening and GNN interpolation in HI-MGN.}
\label{fig:coarsen}
\end{figure}

\paragraph{Coarsening} The hierarchical graphs are constructed by farthest point sampling with Voronoi clustering (hereafter, FPS-Voronoi), as summarized in Algorithm~\ref{alg:fpsvoronoi} and Fig.~\ref{fig:coarsen}. The FPS-Voronoi coarsening requires no externally supplied meshes and each coarse graph is built from the graph of the previous level. First, a user-specified number $k$ of nodes from the fine graph are selected as seeds by greedy FPS~\cite{qi2017pointnet++}.
Second, each fine node is assigned to its nearest seed by a multi-source breadth-first search, where the distance is the number of edges on the shortest path in the fine graph. Nodes assigned to the same seed form a cluster, and we denote the cluster of node $i$ by $C(i)$ and the seed of cluster $c$ by $s_c$. Then, each cluster becomes one coarse node located at its seed, and two coarse nodes are connected by a coarse edge if at least one edge of the fine graph joins their clusters. Therefore, the coarse graph does not generate artificial connections between separate bodies or across the gap of a concave part. Each coarse edge stores the same features as a mesh edge, evaluated between the seed positions of its two end nodes, and each level has its own edge encoder.

\begin{algorithm}[htbp]
\caption{FPS-Voronoi coarsening strategy of HI-MGN.}
\label{alg:fpsvoronoi}
\algrenewcommand\algorithmicrequire{\textbf{Input:}}
\algrenewcommand\algorithmicensure{\textbf{Output:}}

\begin{algorithmic}[1]

\Require Fine graph with node set $V$ and edge set $E$; reference position $p_i$ of each node $i\in V$; number of seeds $k$
\Ensure Seeds $s_1,\dots,s_k$; cluster $C(i)$ of each node $i\in V$; coarse graph with node set $V'$ and edge set $E'$

\Statex \textbf{Step 1: seed selection by FPS}
\State $s_1 \gets$ initial seed node
\State $\rho_i \gets \lVert p_i - p_{s_1} \rVert$ for all $i \in V$ \Comment{$\rho_i$: distance to the nearest seed}
\For{$c = 2,\dots,k$}
    \State $s_c \gets \arg\max_{i \in V} \rho_i$ \Comment{the node farthest from the seeds}
    \State $\rho_i \gets \min\!\big(\rho_i,\,\lVert p_i - p_{s_c} \rVert\big)$ for all $i \in V$
\EndFor

\Statex \textbf{Step 2: Voronoi clustering}
\State Let $\mathrm{hop}(i,j)$ be the number of edges between $i$ and $j$
\State $C(i) \gets \arg\min_{c}\,\mathrm{hop}(i,s_c)$ for all $i \in V$ \Comment{cluster of the nearest seed}

\Statex \textbf{Step 3: coarse graph}
\State $V' \gets \{1,\dots,k\}$ \Comment{one coarse node per cluster, located at its seed}
\ForAll{edges $(i,j) \in E$}
    \If{$C(i) \neq C(j)$} \Comment{the edge joins two different clusters}
        \State $E' \gets E' \cup \{(C(i),C(j))\}$ \Comment{connect the two coarse nodes}
    \EndIf
\EndFor

\end{algorithmic}
\end{algorithm}

\paragraph{Pooling} HI-MGN uses mean pooling for coarsening. Pooling sets the latent state of each coarse node to the mean latent state of the fine nodes in its cluster,
\begin{equation}
  z_c=\frac{1}{n_c}\sum_{i:\,C(i)=c}z_i,
  \label{eq:meanpool}
\end{equation}
where $z_c$ is the latent state of coarse node $c$, $z_i$ is the latent state of fine node $i$, and $n_c$ is the number of fine nodes in cluster $c$.

\paragraph{Interpolation} The interpolation lifts the coarse latent states back to the fine level using a learned GNN, which we refer to as the interpolation GNN as shown in Fig.~\ref{fig:coarsen}. Each fine node $i$ receives messages from the coarse node of its own cluster and the connected coarse nodes. We denote these coarse nodes as the source nodes of node $i$. Each source node $c$ sends node $i$ a message $\mu_{ic}$ built from the coarse latent state $z_c$, the skip state $z^{\mathrm{skip}}_i$, and the relative position $p_i-p_{s_c}$. Thus, the message can be expressed as
\begin{equation}
  \mu_{ic} = \psi([z_c,z^{\mathrm{skip}}_i,p_i-p_{s_c}]),
  \label{eq:interp_msg}
\end{equation}
where $\psi$ is an MLP. The messages are then combined with the skip state by a second MLP $\psi_{V}$, which gives the interpolated latent state $z^{\mathrm{up}}_i$,
\begin{equation}
  z^{\mathrm{up}}_i
    = \psi_{V}([z^{\mathrm{skip}}_i,\sum_{c}\mu_{ic}]),
  \label{eq:interp_node}
\end{equation}
where the sum is taken over the source nodes $c$ of node $i$.
Finally, $z^{\mathrm{up}}_i$ is merged with the skip state by a trainable linear layer, which maps $[z^{\mathrm{skip}}_i,z^{\mathrm{up}}_i]$ to the latent state $z_i$ passed to the message passing blocks of the fine level. Each level has its own $\psi$, $\psi_{V}$, and linear layer.

\subsection{Loss and training}
\label{subsec:training}

Input channels, positional features, edge features, and the state changes of Eq.~\eqref{eq:target} are normalized to zero mean and unit variance. The network minimizes a channel-weighted mean squared error on the normalized physical-state updates,
\begin{equation}
  \mathcal{L} = \frac{1}{N}\sum_{i}\sum_{q} w_q\,
  \big(\hat{y}_{i,q}-y_{i,q}\big)^{2},
  \label{eq:loss}
\end{equation}
where $\hat{y}_{i,q}$ is the decoder prediction for output channel $q$ of node $i$, $y_{i,q}$ is the corresponding normalized state change, $N$ is the number of nodes, and the channel weights $w_q$ sum to one.

Following the noise-injection strategy used in MeshGraphNets \cite{pfaff2020learning}, zero-mean Gaussian noise is added to the temporal physical-state inputs during training of transient problems, and the prediction targets are corrected consistently with the applied perturbation to improve autoregressive rollout robustness,
\begin{align}
  \tilde{\mathbf{u}}^{t}_{i} &{}=\mathbf{u}^{t}_{i}+\boldsymbol{\epsilon}_{i},
  \label{eq:noise}\\
  \Delta\tilde{\mathbf{u}}^{t}_{i} &{}=\mathbf{u}^{t+1}_{i}-\tilde{\mathbf{u}}^{t}_{i},
  \label{eq:noise_target}
\end{align}
where $\tilde{\mathbf{u}}^{t}_{i}$ is the perturbed input, $\Delta\tilde{\mathbf{u}}^{t}_{i}$ is the corrected target, and each component of the noise $\boldsymbol{\epsilon}_{i}$ has a standard deviation of $\sigma$ in normalized units. Edge features are computed from the unperturbed state. In static problems, where the solution is predicted in a single step, the model is trained without input noise.

\section{Numerical examples}\label{sec:examples}

\paragraph{Compared models} In this section, we compare the results of HI-MGN with those of four baseline GNN models: MeshGraphNets \cite{pfaff2020learning}, BSMS-GNN \cite{cao2023efficient}, ROBIN \cite{wurth2025diffusion}, and the Vanilla GNN of DGN \cite{lino2025learning}. These models are chosen because they are representative of modern hierarchical GNNs and share some of the characteristics summarized in Table~\ref{tab:hier_props}. In particular, ROBIN was reported to achieve state-of-the-art GNN accuracy on nonlinear solid mechanics benchmarks including contact \cite{wurth2025diffusion}. The Vanilla GNN is the deterministic counterpart of DGN, which has the same multiscale architecture as DGN but predicts in a single forward pass without diffusion \cite{lino2025learning}. All of the methods mentioned above are implemented using their authors' code with minor modifications and verified against the results from the original papers, except for MeshGraphNets which shares the code base with HI-MGN when hierarhcy is not used.


\paragraph{Training} For the considerd models, all use the AdamW optimizer with gradient clipping, learning-rate warmup followed by cosine annealing, mixed-precision training, and an exponential moving average (EMA) of the model weights. The values of these settings are listed in \ref{app:details}, while the input and output channels, number of training epochs, and input noise are given for each example below. All predictions are made with the EMA model, and the reported accuracies are computed on separate test sets that are never used in training.

\paragraph{Computational cost} Unless stated otherwise, each model is trained on a single NVIDIA GeForce RTX 3090 GPU with Intel Xeon Gold 5218R CPUs. For each model, we report the number of trainable parameters, peak VRAM usage, training time, and inference time of one prediction, measured as described in \ref{app:details}.

\paragraph{Accuracy metric} The accuracy of each predicted field is measured by the coefficient of determination over the nodes, written as
\begin{equation}
  R^2 = 1 - \frac{\sum_i (\hat{f}_i - f_i)^2}{\sum_i (f_i - \bar{f})^2},
  \label{eq:r2}
\end{equation}
where $f_i$ and $\hat{f}_i$ are the reference and predicted values at node $i$, and $\bar{f}$ is the mean of the reference values. The mean $R^2$ in the tables is the average over the fields of a benchmark. In the following tables showing accuracy, the best value is shown in bold and underlined, and the second best is underlined.

\subsection{3D structural contact analysis}\label{subsec:contact}

\begin{figure}[h]
    \centering
    \includegraphics[width=1.0\linewidth]{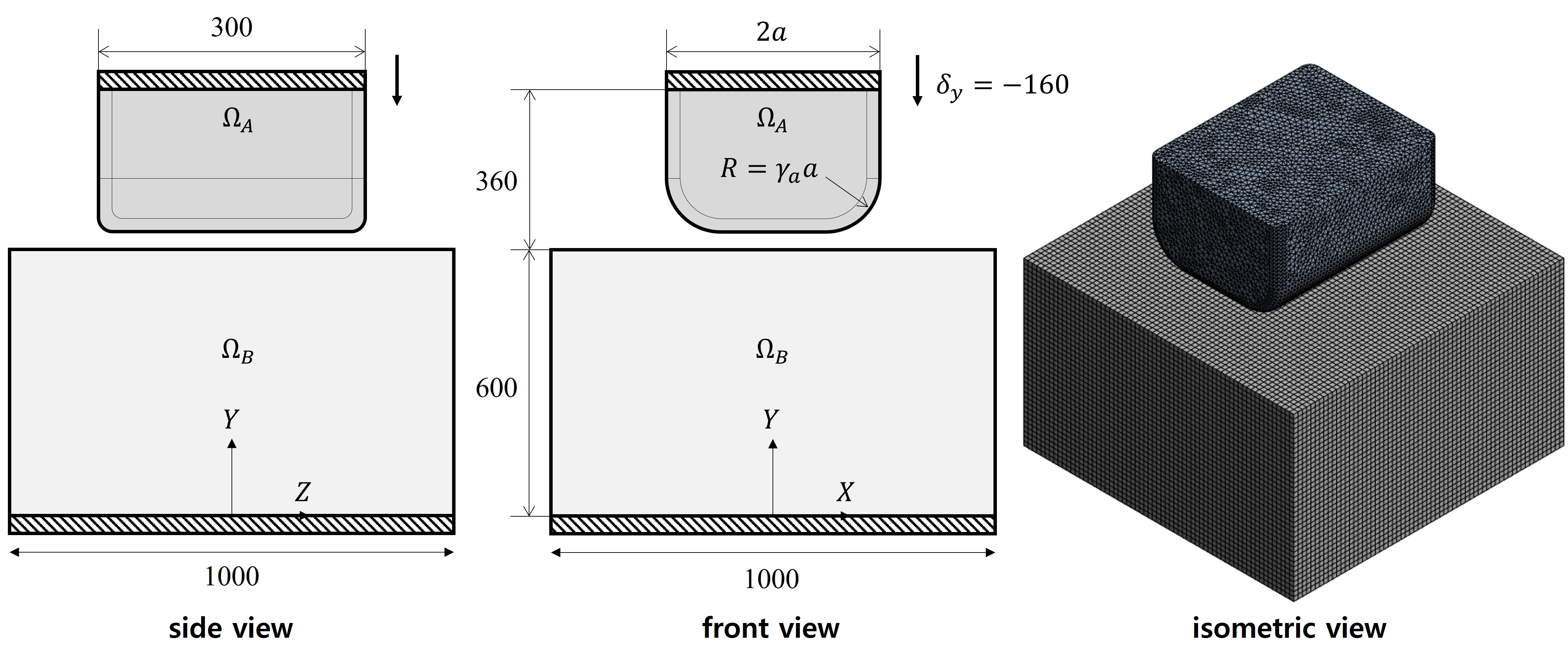}
    \caption{Schematic of the contact benchmark.}
    \label{fig:contact_schematics}
\end{figure}

First, we consider a three-dimensional structural contact problem. Fig.~\ref{fig:contact_schematics} illustrates the configuration, where the indenter ($\Omega_A$) moves downward to press the rectangular parallelepiped foundation block ($\Omega_B$). With the coordinate origin placed at the center of the bottom face of the foundation block, the configuration is symmetric with respect to the $XY$- and $YZ$-planes. The top surface of the indenter is prescribed with a displacement of $\delta_y=-160\,\mathrm{mm}$, which is applied over 50 steps, while the bottom surface of the foundation block is fixed. The indenter and foundation block, designated as the contact and target bodies, are discretized with 4-node tetrahedral elements and 8-node hexahedral elements, respectively. The indenter dimensions are parameterized and vary across samples, with a width of $2a$ and a contact surface radius of $R=\gamma_a a$, where the two parameters
\[
  a\in[160,\,320]\,\mathrm{mm},\qquad \gamma_a\in[0.125,\,0.875]
\]
are selected via Latin hypercube sampling with 50 samples. In Fig.~\ref{fig:contact_schematics}, the isometric view shows the finite element discretization for $a=240\,\mathrm{mm}$ and $\gamma_a=0.5$. Both bodies consist of a compressible neo-Hookean hyperelastic material, whose properties are summarized in Table~\ref{tab:contact_properties}.

\begin{table}[h]
    \centering
    \begin{tabular}{c|cc}
        \hline
        Property & Indenter ($\Omega_A$) & Foundation block ($\Omega_B$)\\
        \hline
        Elastic modulus              & $100\ \mathrm{Pa}$ & $1\ \mathrm{Pa}$ \\
        Poisson's ratio              & $0.3$ & $0.3$ \\
        \hline
    \end{tabular}
    \caption{Material properties of the indenter and the foundation block.}
    \label{tab:contact_properties}
\end{table}

In this benchmark, the nodal state consists of x-, y-, z-direction displacement components and von Mises stress. The 50 samples are split into 40 training and 10 validation samples, with the same split used for all models, and all models are trained for 100 epochs with input noise of $\sigma=0.01$ for time-stepping stability. For this dataset, we consider a two-level hierarchy ($H=2$) of $[N, 5{,}000, 100]$ for HI-MGN, where $N$ is the number of mesh nodes, with the message passing block distribution [4, 6, 8, 6, 4]. BSMS-GNN uses a six-level hierarchy with 13 message passing blocks, while MeshGraphNets uses no hierarchy, $[N]$. ROBIN uses a three-level hierarchy with 29 message passing blocks, and the Vanilla GNN uses a five-level hierarchy with 22 message passing blocks. MeshGraphNets is evaluated with 15 message passing blocks, following the original study \cite{pfaff2020learning}, and with 20 blocks, as a 28-block model with the same number of blocks as HI-MGN does not fit in single-GPU memory on this mesh. In the ablation study (Section~\ref{sec:ablation}), the 28-block model results are shown, and the model is trained on two GPUs by splitting its blocks.

\begin{figure}[h]
    \centering
    \includegraphics[width=1.0\linewidth]{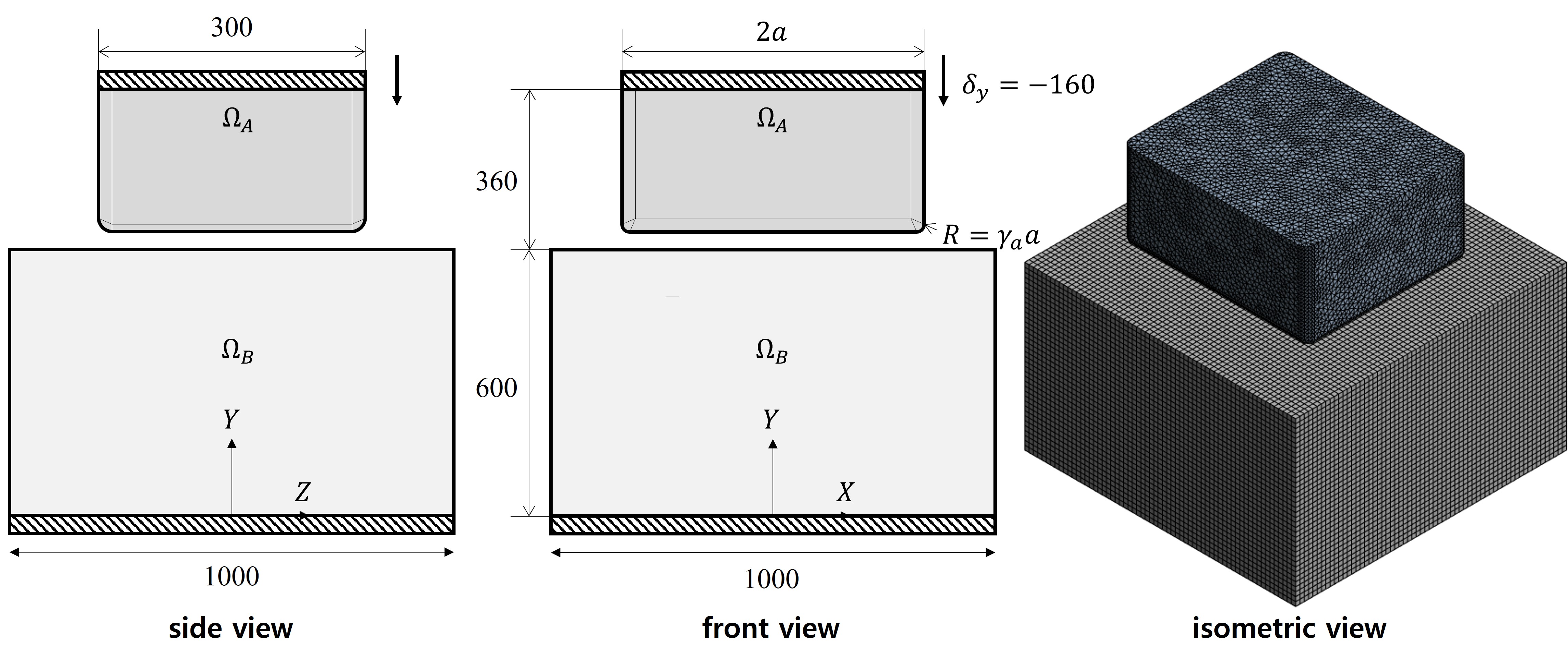}
    \caption{Extrapolation case of the contact benchmark.}
    \label{fig:ex2_infer_schematic}
\end{figure}

\begin{figure}[htbp]
    \centering
    \includegraphics[width=0.78\linewidth]{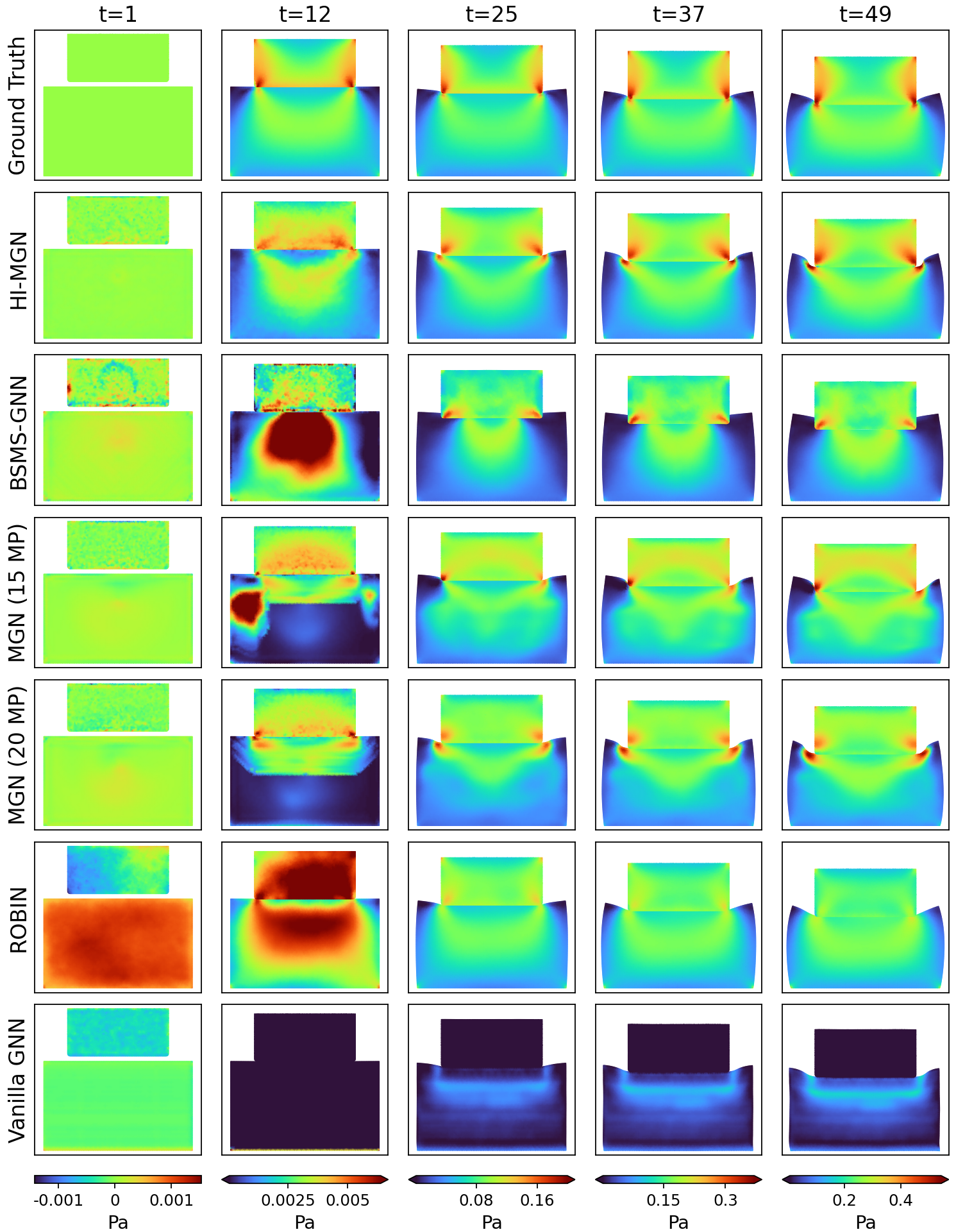}
    \caption{Inferred stress for the extrapolation case.}
    \label{fig:ex2_inference}
\end{figure}

To test each model, we consider both geometric interpolation and extrapolation. The models are evaluated on ten test indenters not included in the training dataset, five of which lie outside the training ranges of $a$ and $\gamma_a$. Details of these test cases are provided in \ref{app:test_cases}. In addition, a large indenter with only a small rounding radius, kept for numerical stability, is selected to test the models on stress concentration with a large spatial gradient in the physics field. This test case, with $a=340$ mm and $\gamma_a=0.05$, is shown in Fig.~\ref{fig:ex2_infer_schematic} together with its finite element discretization. The inference results for this geometric-extrapolation case are shown in Fig.~\ref{fig:ex2_inference}, where von Mises stress on the central section is plotted with one row per model and one column per time step. The color range of each time step follows the ground truth, except at $t=1$, where the ground-truth stress is zero.

\begin{figure}[h]
    \centering
    \includegraphics[width=\linewidth]{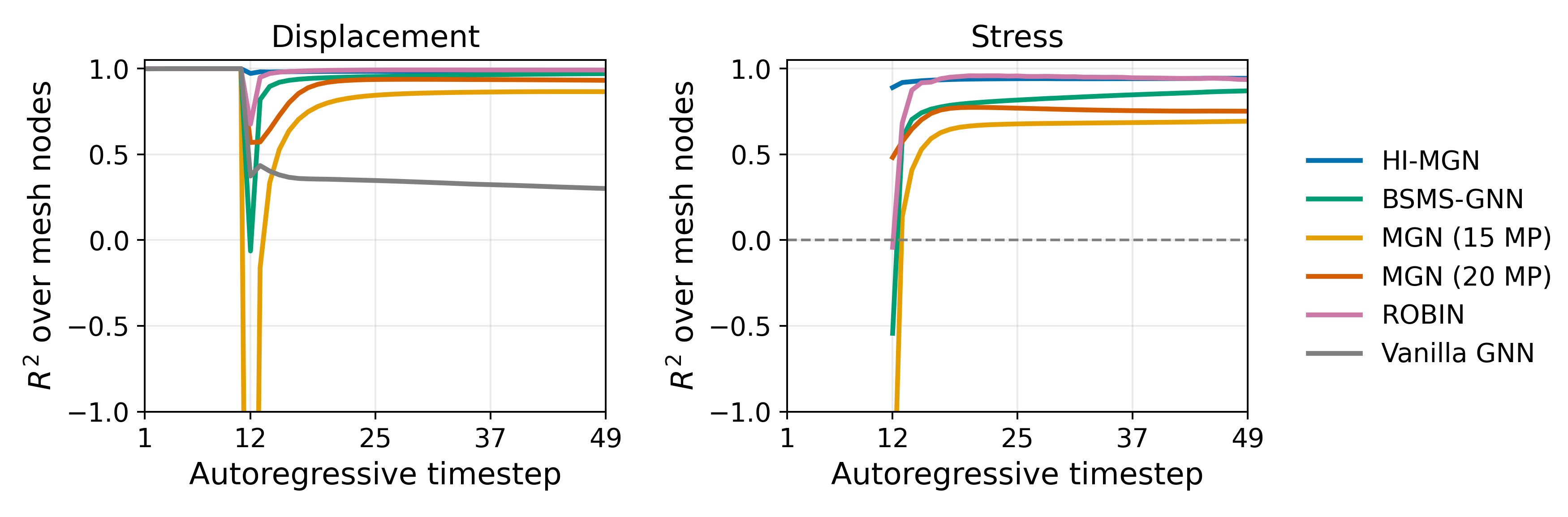}
    \caption{$R^2$ results with respect to time on the contact benchmark.}
    \label{fig:ex2_inference_graph}
\end{figure}

For each of the ten test indenters, $R^2$ is computed at every predicted step and averaged over the steps. Steps before contact onset ($t=12$) are excluded for stress, as it is zero before contact. Table~\ref{tab:contact_r2} lists the time-averaged $R^2$ and Fig.~\ref{fig:ex2_inference_graph} shows $R^2$ at each step, both averaged over the ten indenters.

\begin{table}[h]
    \centering
    \setlength{\tabcolsep}{3pt}
    \begin{tabular}{l|cccccc}
        \hline
        Metric & \hdtwo{HI-MGN}{(proposed)} & \hdtwo{BSMS-}{GNN} & \hdtwo{MGN}{(15MP)} & \hdtwo{MGN}{(20MP)} & ROBIN & \hdtwo{Vanilla}{GNN} \\
        \hline
        Disp. $x$ $R^2$ & \textbf{\underline{0.970}} & 0.877 & 0.271 & 0.825 & \underline{0.967} & $-$0.196 \\
        Disp. $y$ $R^2$ & \underline{0.998} & 0.992 & 0.983 & 0.991 & \textbf{\underline{0.999}} & 0.994 \\
        Disp. $z$ $R^2$ & \textbf{\underline{0.981}} & 0.909 & 0.609 & 0.871 & \underline{0.977} & 0.221 \\
        Stress $R^2$ & \textbf{\underline{0.938}} & 0.784 & 0.586 & 0.743 & \underline{0.913} & $-$4.160 \\
        Mean $R^2$ & \textbf{\underline{0.961}} & 0.855 & 0.603 & 0.819 & \underline{0.947} & $-$1.910 \\
        \hline
    \end{tabular}
    \caption{Time-averaged $R^2$ on the contact benchmark.}
    \label{tab:contact_r2}
\end{table}

Table~\ref{tab:contact_r2} shows that HI-MGN outperforms the existing methods in terms of accuracy. HI-MGN yields the highest mean $R^2$, followed by ROBIN. The largest differences between the methods are observed in Fig.~\ref{fig:ex2_inference} at $t=12$, where contact begins and the stress field develops over both bodies within a single time step. It is observed that HI-MGN accurately predicts the stress concentration at the indenter corners, while the other models fail to capture this feature. Both the 15- and 20-block MeshGraphNets show banded stress patterns below the contact surface, showing the limited long-range information propagation. Such limited information propagation is observed as the receptive field of conventional flat message passing is restricted by the number of message passing blocks. BSMS-GNN also exhibits noticeable errors despite its six-level hierarchy, as it overestimates the stress below the indenter at $t=12$. ROBIN is the closest to HI-MGN in accuracy, as its $R^2$ recovers shortly after contact onset, but it yields a smooth stress field without the stress concentrations at the indenter corners. The $R^2$ at each autoregressive time step is shown in Fig.~\ref{fig:ex2_inference_graph}.

\begin{table}[h]
    \centering
    \small
    \setlength{\tabcolsep}{2.5pt}
    \begin{tabular}{l|cccccc}
        \hline
        Metric & \hdtwo{HI-MGN}{(proposed)} & \hdtwo{BSMS-}{GNN} & \hdtwo{MGN}{(15MP)} & \hdtwo{MGN}{(20MP)} & ROBIN & \hdtwo{Vanilla}{GNN} \\
        \hline
        VRAM (GB) & \textbf{\underline{15.68}} & \underline{17.73} & 19.97 & 24.61 & 24.01 & 21.29 \\
        Train time (hr) & \textbf{\underline{50.46}} & 112.32 & \underline{68.92} & 91.73 & 100.06 & 76.84 \\
        \lbtwo{Inference}{time (ms)} & \underline{990.5} & 1,846.0 & 1,207.0 & 1,371.0 & 14,912.7 & \textbf{\underline{800.9}} \\
        \hline
    \end{tabular}
    \caption{Computational cost on the contact benchmark.}
    \label{tab:contact_eff}
\end{table}

Table~\ref{tab:contact_eff} summarizes the computational costs of the compared models. HI-MGN is found to require the least VRAM and the shortest training time. Its inference time is the second shortest after that of the Vanilla GNN, while the diffusion-based ROBIN, with 20 denoising steps per prediction, requires about 15 times the inference time of HI-MGN.

\subsection{3D steady aerodynamic analysis}\label{subsec:aero}

Next, we consider a 3D steady aerodynamic analysis of the NASA Common Research Model (NASA-CRM). The NASA-CRM is a publicly available benchmark configuration representing a scaled transport aircraft and has been widely recognized in the aerodynamics community. Here, we adopt the simulation results generated by the German Aerospace Center (DLR) using the DLR TAU Reynolds-Averaged Navier--Stokes (RANS) solver \cite{bekemeyer2025introduction}. The dataset consists of steady-state pressure and skin friction results under various flight conditions. Computational fluid dynamics (CFD) results are provided for six varying input parameters, the freestream Mach number ($Ma$), angle of attack (AoA), inboard and outboard aileron deflections, elevator deflection, and horizontal tailplane angle. The original dataset consists of 454,404 surface nodes with the pressure coefficient ($C_p$), and the skin friction coefficients in the $x$, $y$, and $z$ directions ($C_{f_x}, C_{f_y}, C_{f_z}$). Further information regarding the NASA-CRM dataset can be found in the work of Bekemeyer et al. \cite{bekemeyer2025introduction}.

In this section, we train and compare the results on a subsampled version of the original DLR dataset due to the large memory requirements of some models. Based on the DLR results, the original surface nodal data are subsampled to 122,778 nodes with reconstructed edges. The dataset contains 105 flight conditions for training and 44 conditions for testing, and the 105 conditions are split into 84 training and 21 validation conditions for all models. We train the HI-MGN using the same hyperparameters as in the previous numerical example, except for the number of training epochs and input noise. The number of training epochs is set to 500 and input noise is excluded as time-stepping stability is not a concern in this static problem. For the other methods, BSMS-GNN uses an eight-level hierarchy with 17 message passing blocks, while MeshGraphNets is evaluated using 15 and 28 message passing blocks. ROBIN and the Vanilla GNN use the same numbers of levels and message passing blocks as in the previous example.

\begin{figure}[htbp]
    \centering
    \includegraphics[width=\linewidth]{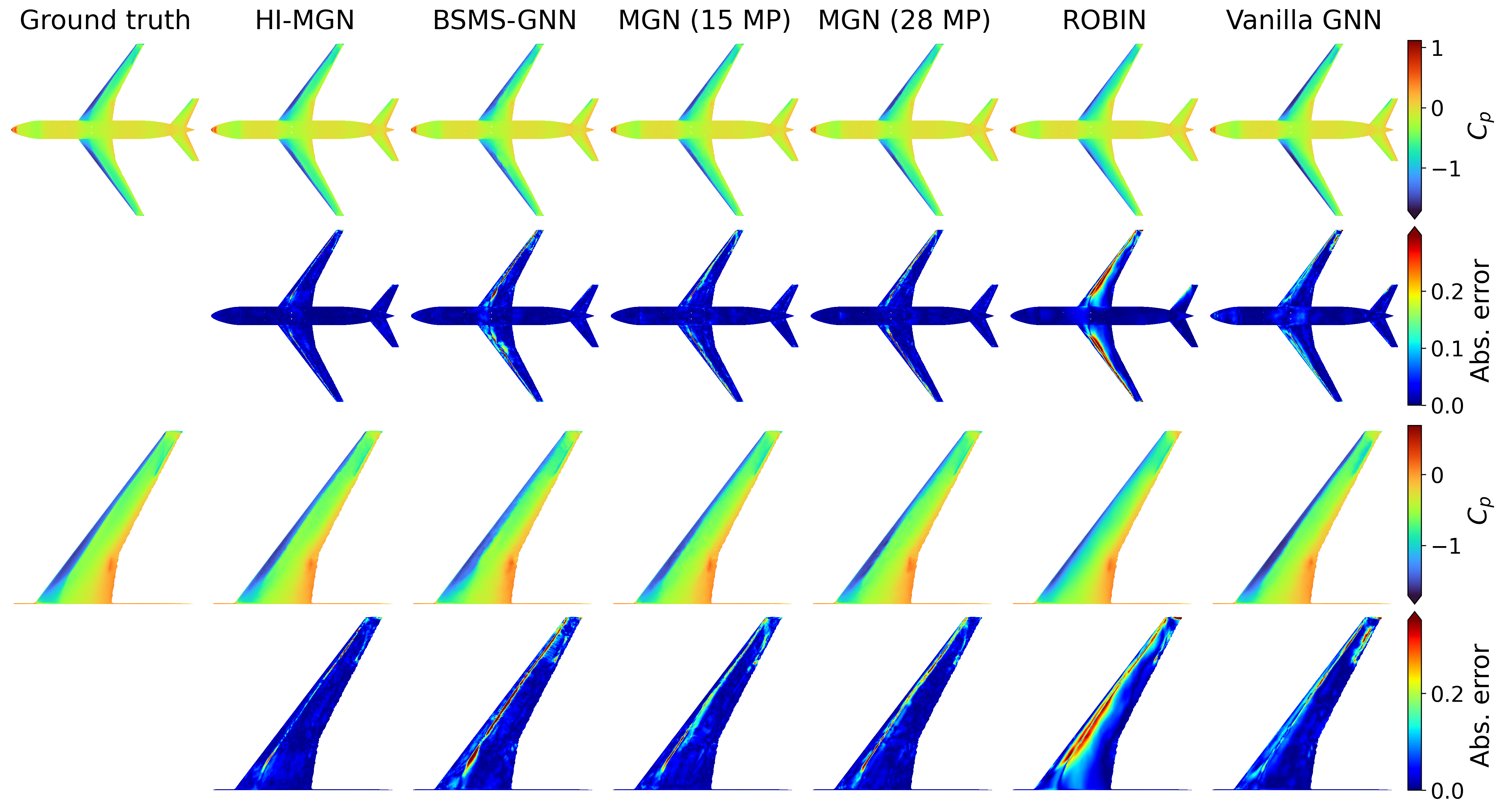}
    \caption{Inferred $C_p$ results for test sample 37.}
    \label{fig:ex3_inference}
\end{figure}

For inference, the compared models are evaluated on the 44 test cases. Fig.~\ref{fig:ex3_inference} shows the predicted $C_p$ for test sample 37, with the absolute error of each model below its prediction, for the whole aircraft and one main wing. This sample corresponds to $Ma$: 0.76, AoA: $3.83^\circ$, inboard ailerons: $-8.22^\circ$, outboard ailerons: $8.85^\circ$, elevator: $1.25^\circ$, and horizontal tailplane: $0.11^\circ$. This case exhibits a clearly visible transonic shock on the upper wing surface and the $R^2$ values averaged over all test samples are also shown in Table~\ref{tab:aero_r2}. 

\begin{table}[h]
    \centering
    \setlength{\tabcolsep}{3pt}
    \begin{tabular}{l|cccccc}
        \hline
        Metric & \hdtwo{HI-MGN}{(proposed)} & \hdtwo{BSMS-}{GNN} & \hdtwo{MGN}{(15MP)} & \hdtwo{MGN}{(28MP)} & ROBIN & \hdtwo{Vanilla}{GNN} \\
        \hline
        $C_p$ $R^2$ & \textbf{\underline{0.987}} & 0.982 & 0.983 & 0.984 & 0.979 & \underline{0.986} \\
        $C_{f_x} R^2$ & \underline{0.966} & 0.960 & 0.962 & 0.962 & 0.957 & \textbf{\underline{0.972}} \\
        $C_{f_y} R^2$ & \textbf{\underline{0.966}} & 0.957 & 0.958 & \underline{0.962} & 0.937 & 0.934 \\
        $C_{f_z} R^2$ & \textbf{\underline{0.994}} & 0.991 & \underline{0.993} & \textbf{\underline{0.994}} & 0.983 & 0.981 \\
        Mean $R^2$ & \textbf{\underline{0.981}} & 0.976 & 0.977 & \underline{0.979} & 0.969 & 0.974 \\
        \hline
    \end{tabular}
    \caption{Sample-averaged $R^2$ on the subsampled NASA-CRM dataset.}
    \label{tab:aero_r2}
\end{table}

\begin{table}[h]
    \centering
    \small
    \setlength{\tabcolsep}{2.5pt}
    \begin{tabular}{l|cccccc}
        \hline
        Metric & \hdtwo{HI-MGN}{(proposed)} & \hdtwo{BSMS-}{GNN} & \hdtwo{MGN}{(15MP)} & \hdtwo{MGN}{(28MP)} & ROBIN & \hdtwo{Vanilla}{GNN} \\
        \hline
        VRAM (GB) & \underline{7.42} & \textbf{\underline{3.68}} & 10.54 & 16.80 & 11.91 & 10.74 \\
        Train time (hr) & \underline{6.22} & \textbf{\underline{5.59}} & 8.18 & 14.83 & 11.78 & 6.48 \\
        \lbtwo{Inference}{time (ms)} & 433.5 & \underline{402.0} & 543.5 & 852.5 & 7,050.9 & \textbf{\underline{228.9}} \\
        \hline
    \end{tabular}
    \caption{Computational cost on the subsampled NASA-CRM dataset.}
    \label{tab:aero_eff}
\end{table}

The results shown in Fig.~\ref{fig:ex3_inference} and Table~\ref{tab:aero_r2} indicate that all compared models achieve a relatively high prediction accuracy with minor differences. HI-MGN achieves the highest $R^2$ values for $C_p$, $C_{f_y}$, $C_{f_z}$, and the highest mean $R^2$. For $C_{f_x}$, the Vanilla GNN is the most accurate, followed by HI-MGN.

In contrast, the computational cost differences are more pronounced. As summarized in Table~\ref{tab:aero_eff}, BSMS-GNN has the lowest peak VRAM usage and shortest training time. This is because the dataset consists of only the 3D surface mesh, making it topologically similar to a 2D mesh, so that the number of bi-stride edges decreases with every level unlike the volumetric 3D mesh. HI-MGN ranks second in both the VRAM and training time. Its inference time is the third shortest, while ROBIN requires about 16 times the inference time of HI-MGN.

\subsection{Full NASA-CRM dataset and comparison with existing methods}\label{subsec:crm_full}

\begin{figure}[htbp]
    \centering
    \begin{minipage}[t]{0.49\linewidth}
        \centering
        \includegraphics[width=\linewidth]{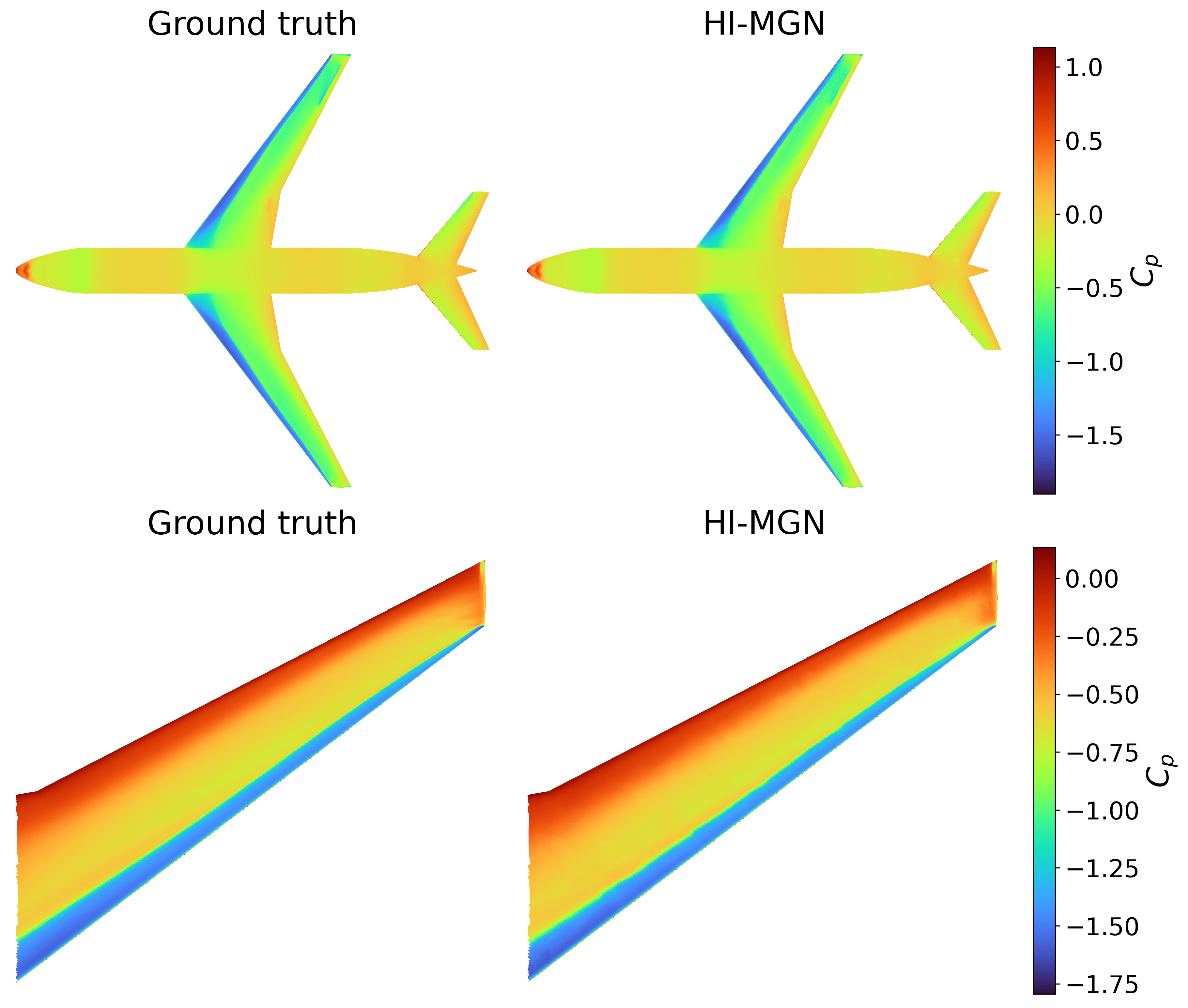}\\
        {\small(a) Test sample 37}
    \end{minipage}\hfill
    \begin{minipage}[t]{0.49\linewidth}
        \centering
        \includegraphics[width=\linewidth]{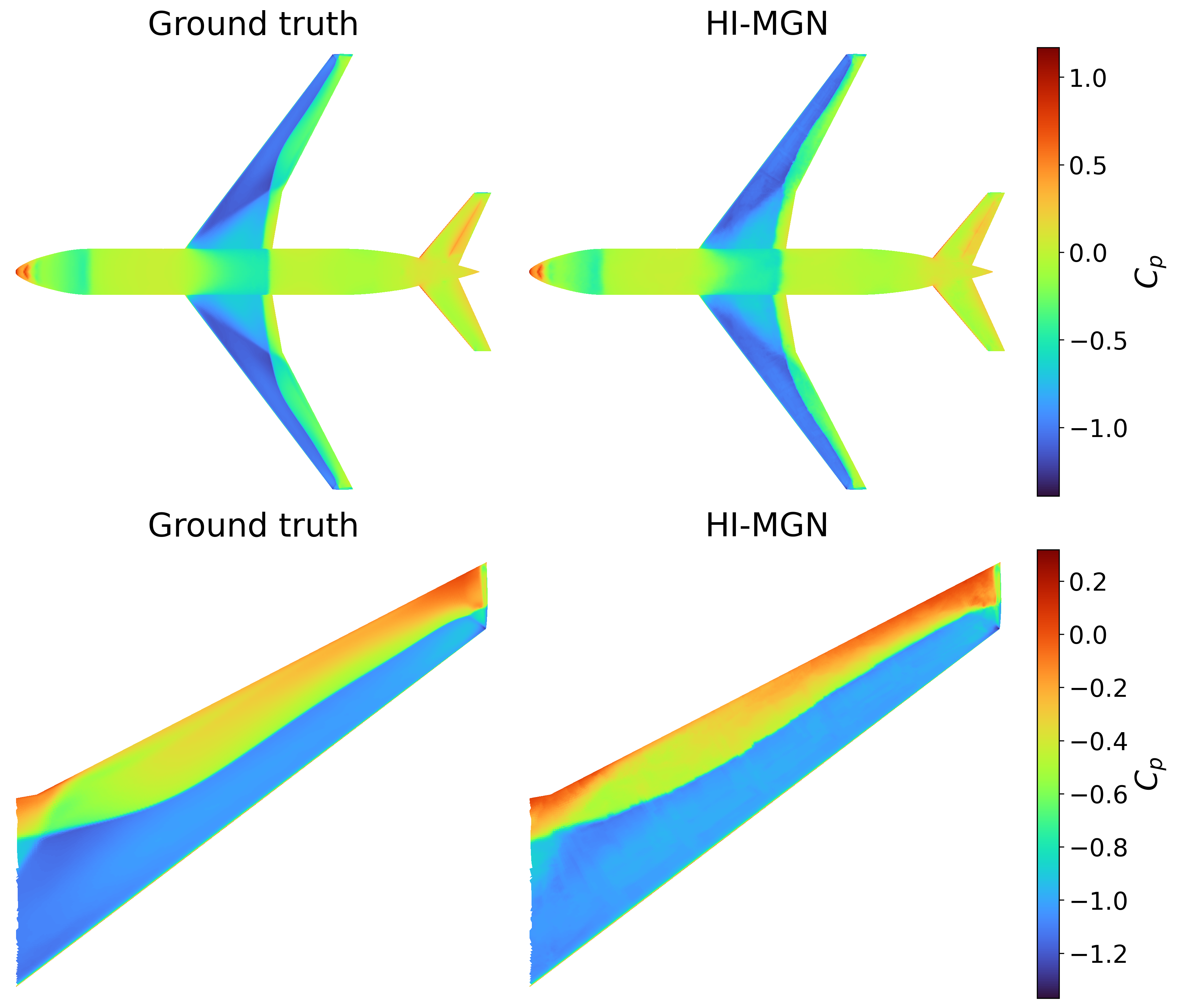}\\
        {\small(b) Test sample 31}
    \end{minipage}
    \caption{Inferred $C_p$ results on the full NASA-CRM dataset.}
    \label{fig:NASA-CRM_inference}
\end{figure}

HI-MGN can also be trained on the full NASA-CRM dataset, i.e., the original dataset of 454,404 nodes without subsampling, using a consumer-grade hardware. We use the same hierarchy, message passing blocks, and hyperparameters as in Section~\ref{subsec:aero}, but the model is trained on all 105 training conditions and evaluated on the 44 test conditions as other models are. The training is performed for 4,000 epochs, and the training cost and test accuracy are shown in Table~\ref{tab:NASA-CRM_training}, with the predicted $C_p$ shown in Fig.~\ref{fig:NASA-CRM_inference}.

\begin{table}[h]
\centering
\begin{tabular}{c|c|c|c}
\hline
Metric & Value & Metric & Value \\
\hline
VRAM (GB) & 17.63 & Train time (hr) & 156.34 \\
$C_p$ $R^2$ & 0.9886 & $C_{f_x}$ $R^2$ & 0.9861\\
$C_{f_y}$ $R^2$ & 0.9787 & $C_{f_z}$ $R^2$ & 0.9933\\
Mean $R^2$ & 0.9873 & & \\
\hline
\end{tabular}%
\caption{Results of HI-MGN on the full NASA-CRM dataset.}
\label{tab:NASA-CRM_training}
\end{table}

Compared to the subsampled dataset in Section~\ref{subsec:aero}, peak VRAM usage and training time in Table~\ref{tab:NASA-CRM_training} are considerably larger. The full-mesh dataset has about 3.7 times as many nodes and the model is trained for 4,000 epochs instead of 500, while the mean $R^2$ increases from 0.981 to 0.987. Fig.~\ref{fig:NASA-CRM_inference} shows that HI-MGN is capable of predicting transonic shock patterns albeit with small spurious fluctuations. As in Fig.~\ref{fig:ex3_inference}, the whole aircraft is shown at the top and one main wing at the bottom. Fig.~\ref{fig:NASA-CRM_inference}(a) shows test sample 37, the case of Fig.~\ref{fig:ex3_inference}, and Fig.~\ref{fig:NASA-CRM_inference}(b) shows test sample 31 at a higher Mach number ($Ma$: 0.88, AoA: $4.38^\circ$). 

Because the NASA-CRM dataset is publicly available, we compare HI-MGN's performance to that reported by other methods. Table~\ref{tab:NASA-CRM_comparison} lists the relative $L_2$ errors of various methods on the NASA-CRM dataset, where the values for the other methods are taken from the Transolver-3 paper \cite{zhou2026transolver}. These values were obtained on the same 105 training and 44 test conditions, but with that paper's training settings. Following that paper, the error is computed for each test condition and then averaged, with $C_f$ taken as the skin-friction vector magnitude, and the mean is the average of the $C_f$ and $C_p$ errors. HI-MGN achieves the lowest $C_f$ error, the second-lowest $C_p$ error after Transolver-3, and the lowest mean error overall.

\begin{table}[h]
\centering
\begin{tabular}{c|ccc}
\hline
Methods & $C_f$ & $C_p$ & Mean \\
\hline
HI-MGN (proposed) & \textbf{\underline{5.17\%}} & \underline{9.20\%} & \textbf{\underline{7.18\%}} \\
Transolver-3 \cite{zhou2026transolver} & \underline{5.85\%} & \textbf{\underline{8.71\%}} & \underline{7.28\%} \\
AB-UPT \cite{alkin2025ab} & 6.43\% & 9.77\% & 8.10\% \\
Transolver++ \cite{luo2025transolver++} & 6.95\% & 9.51\% & 8.23\% \\
Transolver \cite{wu2024transolver} & 7.04\% & 9.61\% & 8.33\% \\
GINO \cite{li2023geometry} & 11.51\% & 12.39\% & 11.95\% \\
MeshGraphNets \cite{pfaff2020learning} & 9.81\% & 14.80\% & 12.31\% \\
Graph U-Nets \cite{gao2019graph} & 15.61\% & 15.85\% & 15.73\% \\
UPT \cite{alkin2024universal} & 23.78\% & 12.78\% & 18.28\% \\
GAOT \cite{wen2025geometry} & 59.79\% & 30.38\% & 45.09\% \\
\hline
\end{tabular}%
\caption{Relative $L_2$ errors on the NASA-CRM dataset.}
\label{tab:NASA-CRM_comparison}
\end{table}

\section{Ablation study}\label{sec:ablation}
We perform the ablation study on the contact benchmark, where the accuracy gap between HI-MGN and other models are largest. All variants share the same encoder, decoder, training schedule, data split, and loss. The reference model uses a two-level hierarchy $[N, 5{,}000, 100]$, a coarse-centric message passing block distribution of $[4,6,8,6,4]$, FPS-Voronoi latent mean pooling, and a learned interpolation GNN for upsampling. Each block of rows in Table~\ref{tab:ablation} varies one of these choices. The second row removes the hierarchy ($H=0$) while keeping 28 message passing blocks, and the third row removes the positional features of Eq.~\eqref{eq:posfeat}. In the variants with a different hierarchy or number of blocks, the blocks are distributed as $[4,20,4]$ (one-level), $[4,4,4,4,4,4,4]$ (three-level), $[2,3,4,3,2]$ (14 blocks), and $[5,8,12,8,5]$ (38 blocks). In the FPS-latent inherit variant, each coarse node takes the latent state of its FPS seed instead of the mean over its Voronoi cluster. In the linear interpolator variant, no interpolation GNN is used, and each fine node takes the coarse latent state of its own cluster, which is merged with the skip state by the same linear layer as in the reference model. The last two variants keep the interpolation GNN. In the own-cluster variant, each fine node receives a message only from its own cluster, and in the mean-aggregation variant, the messages in Eq.~\eqref{eq:interp_node} are averaged instead of summed.

\begin{table}[!htbp]
\centering
\setlength{\tabcolsep}{4pt}
\begin{tabular}{c|ccc}
\hline
 & \multicolumn{3}{c}{$R^2$} \\ \cline{2-4} Variant & Stress & Displacement & Mean \\
\hline
 Reference & \textbf{\underline{0.9383}} & 0.9831 & \textbf{\underline{0.9607}} \\
\hline
 No hierarchy ($H=0$), 28 blocks$^{*}$ & 0.8470 & 0.9295 & 0.8883 \\
\hline
 No positional features & 0.8628 & 0.9741 & 0.9184 \\
\hline
 One-level: [$N$, 100] & 0.8574 & 0.9732 & 0.9153 \\
Three-level: [$N$, 10,000, 1,000, 100] & 0.9328 & 0.9830 & \underline{0.9579} \\
\hline
 Uniform block distribution: {[5,6,6,6,5]}
    & \underline{0.9353} & 0.9799 & 0.9576 \\
Fine-centric block distribution: {[7,5,4,5,7]}
    & 0.9278 & 0.9830 & 0.9554 \\
\hline
Number of message passing blocks: 14
    & 0.9228 & \underline{0.9847} & 0.9537 \\
Number of message passing blocks: 38
    & 0.9282 & 0.9815 & 0.9548 \\
\hline
FPS-latent inherit downsampling
    & 0.8991 & 0.9711 & 0.9351 \\
\hline
Linear interpolator upsampling
    & 0.9264 & \textbf{\underline{0.9854}} & 0.9559\\
Own-cluster learned upsampling
    & 0.9263 & 0.9778 & 0.9520 \\
Mean-aggregated learned upsampling
    & 0.9230 & 0.9816 & 0.9523 \\
\hline
\end{tabular}%

\vspace{2pt}
{\footnotesize $^{*}$Trained on two GPUs with the blocks split between them (model splitting).}
\caption{Ablation study on the contact benchmark: time-averaged $R^2$, averaged over the ten test indenters.}
\label{tab:ablation}
\end{table}

The results shown in Table \ref{tab:ablation} indicate that the hierarchy has the largest effect on accuracy, as the flat MeshGraphNets yields a much lower $R^2$ than the reference model. The results imply that the largest contributing factor after hierarchy is the number of hierarchy levels, followed by the positional features and the downsampling scheme. The latter indicates that the mean over the Voronoi cluster represents the cluster better than the latent state of the seed alone.

The number of hierarchy levels also affects the accuracy, as the reference two-level hierarchy is the most accurate while the one-level hierarchy is less accurate and adding a third level does not improve the accuracy. Therefore, a deeper hierarchy is not necessarily more accurate, which is suspected to be correlated with the errors introduced by each additional pooling and interpolation. For the block distribution, the coarse-centric distribution yields the most accurate results, as the mean $R^2$ decreases when the blocks are moved toward the input graph. This suggests that message passing on the coarse levels is needed to propagate information over the whole mesh, and coarse to fine interpolation is capable of handling fine details well. In addition, both 14 and 38 blocks yield a lower mean $R^2$ than 28 blocks, and the interpolation of the reference model yields the highest mean $R^2$ among the four interpolation schemes. Overall, the reference model yields the highest mean $R^2$ in Table~\ref{tab:ablation}.

\section{Conclusions}\label{sec:conclusions}

In this paper, we present a new hierarchical extension of MeshGraphNets. The proposed model, HI-MGN, is designed to achieve higher accuracy at a lower computational cost than the existing mesh-based graph neural networks even for large graphs. The developed HI-MGN is a self-contained graph neural network that performs hierarchical message passing across multiple graph resolutions. The model selects coarse nodes using FPS and assigns fine nodes to them through graph-based Voronoi partitioning. It constructs coarse edges from the connectivity of the original mesh, avoiding artificial connections between disconnected components. Upsampling from the coarse to the fine level is performed by a learned interpolation GNN that combines neighboring coarse latent states, fine-level skip states, and relative geometric information to reconstruct the fine-level states.

The results from HI-MGN is compared against that of MeshGraphNets, BSMS-GNN, ROBIN, and the Vanilla GNN across structural contact and steady aerodynamic problems. The numerical results suggest that the advantages of hierarchical message passing are particularly pronounced when the prediction requires information to propagate over large graph distances. On the contact benchmark, the performance differences between HI-MGN and other models are clearly visible, particularly shortly after the onset of contact, when newly generated local contact information must propagate rapidly through the domain. On the NASA-CRM benchmark, the accuracies of all models are close to each other, but still, HI-MGN yields the highest mean $R^2$ on the NASA-CRM benchmark. Not only does HI-MGN exhibit high accuracy, it also is efficient as well. It requires the least VRAM and training time on the contact benchmark, and the second least and second shortest on the NASA-CRM benchmark. ROBIN, the second accurate model on the contact benchmark, requires about 15 times the inference time of HI-MGN due to its diffusion process. In addition, as HI-MGN is efficient in terms of VRAM, it can be trained on the full NASA-CRM dataset with a single consumer-grade GPU. The results on the full NASA-CRM dataset yields the lowest mean error compared to other state-of-the-art models such as Transolver-3 and AB-UPT. While this is not a definitive result suggesting HI-MGN is the state-of-the-art model, it indicates that HI-MGN is comparable to these in terms of accuracy.

Finally, the ablation study on the contact benchmark confirms that removing the hierarchy substantially reduces accuracy, with the mean $R^2$ dropping from 0.961 to 0.888. Among the other design choices, the number of hierarchy levels, the positional features, and the pooling and interpolation schemes have the largest effects on accuracy. Overall, the results suggest that hierarchical topology-aware message passing provides an effective approach for mitigating the long-range interaction and computational and memory limitations of conventional mesh-based GNNs. Rather than increasing the depth of a flat processor, HI-MGN reduces graph interaction distances through coarsening and performs a large portion of the message passing on substantially smaller graphs, which reduces the number of latent node and edge states on the original mesh that must be retained during training. These characteristics make HI-MGN a practical framework for unstructured engineering meshes involving complex geometries, multiple components, deformation, and long-range physical interactions. In future work, we will consider the automatic selection of the optimal number of hierarchy levels and coarse nodes, which are currently chosen by the user.

\section*{Declaration of competing interest}
The authors declare that they have no known competing financial interests or personal relationships that could have appeared to influence the work reported in this paper.

\section*{Data availability}
The source code for HI-MGN, including the graph-coarsening and interpolation modules, training configurations, benchmark preprocessing scripts, and scripts used to reproduce the main results of this study, is publicly available at https://github.com/leesihun/HI-MGN. The finite element dataset of the 3D contact benchmark, which was generated for this study, is available from the corresponding author on reasonable request. The NASA-CRM data are those of the surrogate modeling benchmark described by Bekemeyer et al. \cite{bekemeyer2025introduction}.

\appendix
\setcounter{table}{0}
\section{Training and dataset details}\label{app:details}

Table~\ref{tab:train_settings} lists the training settings used to train the considered models. All models are trained with gradient checkpointing and mixed precision to reduce VRAM usage. We also implement an EMA model for inference. Peak VRAM usage is the largest GPU memory usage during training, in units of $10^9$ bytes. Training time is the wall-clock time from the start of the preprocessing to the end of the training. The reported inference time is the median wall-clock time of one prediction step over the test samples.

\begin{table}[htbp]
    \centering
    \begin{tabular}{ll}
        \hline
        Setting & Value \\
        \hline
        Optimizer & AdamW \\
        Learning rate & $10^{-4}$ with linear warmup and cosine annealing\\
        Warmup epochs & 3 \\
        Gradient norm clipping & 3.0 \\
        EMA decay & 0.99 \\
        Batch size & 1 \\
        Latent dimension $d_z$ & 128 \\
        MLP hidden width & 128 \\
        \hline
    \end{tabular}
    \caption{Hyperparameters used to train the considered models.}
    \label{tab:train_settings}
\end{table}

Table~\ref{tab:mesh_stats} summarizes the dataset mesh sizes, with the minimum and maximum over the samples for the contact meshes. Edges are counted once per node pair. The diameter is the largest number of edges on a shortest path between any two nodes in a graph. The subsampled NASA-CRM dataset uses reconstructed edges, and all flight conditions share the same mesh.

\begin{table}[htbp]
    \centering
    \setlength{\tabcolsep}{3pt}
    \begin{tabular}{llcccc}
        \hline
        Dataset & Split & Samples & Nodes & Edges & \hdtwo{Diameter}{(hops)} \\
        \hline
        3D contact & Training & 50 & 183,487-- & 652,840-- & 109 \\
         & & & 206,087\phantom{--} & 804,436\phantom{--} & \\[3pt]
         & Test & 10 & 177,346-- & 611,941-- & 109 \\
         & & & 214,644\phantom{--} & 861,510\phantom{--} & \\
        \hline
        Subsampled NASA-CRM & Training & 105 & 122,778\phantom{--} & 405,292\phantom{--} & 527 \\
         & Test & 44 & 122,778\phantom{--} & 405,292\phantom{--} & 527 \\
        \hline
    \end{tabular}
    \caption{Details of the 3D contact and subsampled NASA-CRM meshes.}
    \label{tab:mesh_stats}
\end{table}

Table~\ref{tab:level_stats} lists the numbers of nodes and edges, the mean degree (the mean number of neighbors per node, $2|E|/|V|$), and the diameter on each level of the HI-MGN hierarchy built on the largest mesh of each benchmark, where level 0 is the input graph. For the contact benchmark, the diameter is that of the block.

\begin{table}[htbp]
    \centering
    \small\setlength{\tabcolsep}{3pt}\begin{tabular}{l|cccc|cccc}
        \hline
         & \multicolumn{4}{c|}{3D contact} & \multicolumn{4}{c}{NASA-CRM} \\
        Level & Nodes & Edges & Degree & \hdtwo{Diameter}{(hops)} & Nodes & Edges & Degree & \hdtwo{Diameter}{(hops)} \\
        \hline
        0 & 214,644 & 861,510 & 8.03 & 109 & 122,778 & 405,292 & 6.60 & 527 \\
        1 & 5,000 & 30,973 & 12.39 & 23 & 5,000 & 15,116 & 6.05 & 95 \\
        2 & 100 & 474 & 9.48 & 5 & 100 & 272 & 5.44 & 16 \\
        \hline
    \end{tabular}
    \caption{Sizes of the HI-MGN hierarchy levels.}
    \label{tab:level_stats}
\end{table}

\setcounter{table}{0}
\setcounter{figure}{0}
\section{Test case details for the numerical examples}\label{app:test_cases}

\begin{figure}[htbp]
    \centering
    \includegraphics[width=0.8\linewidth]{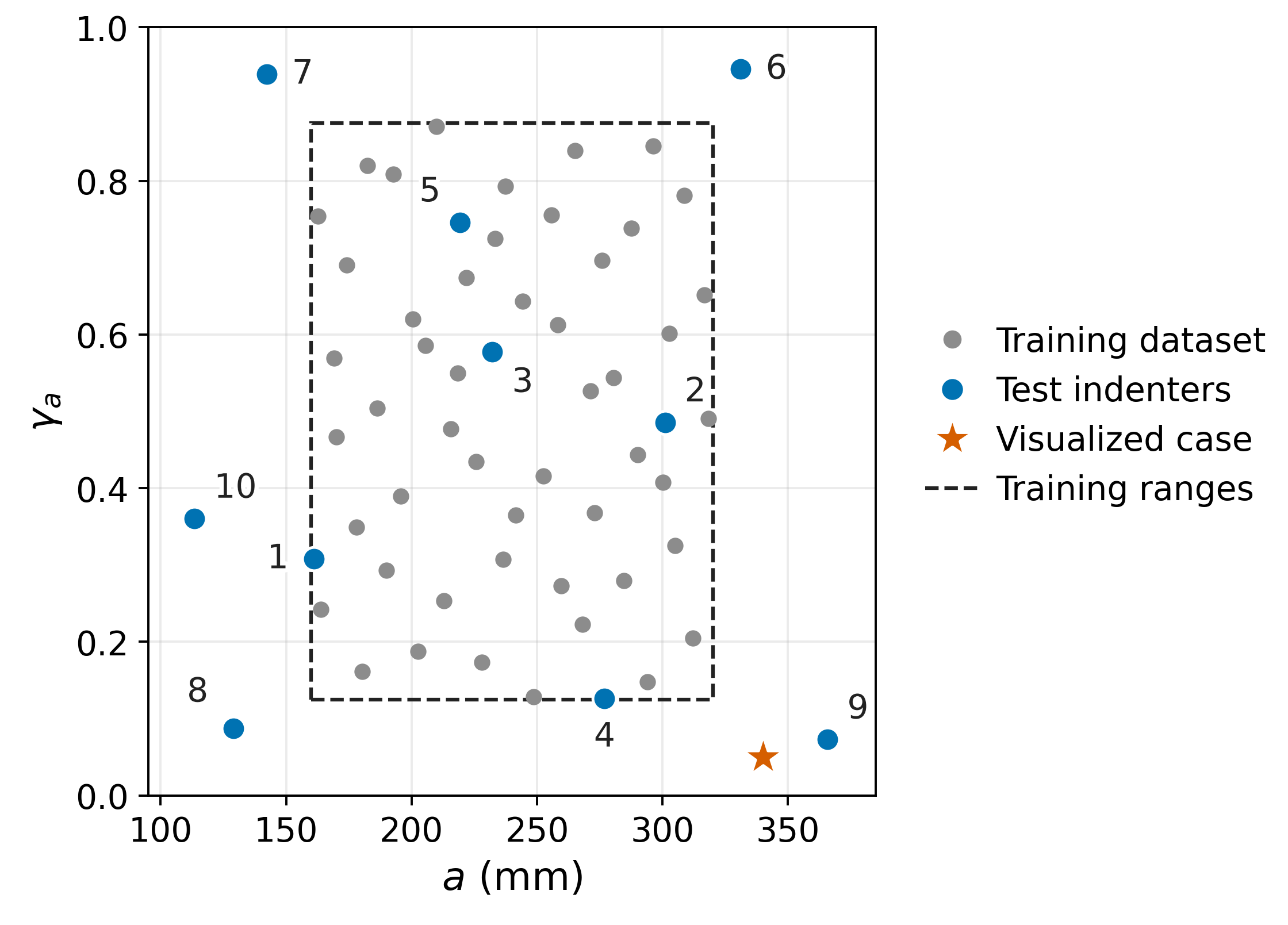}
    \caption{Indenter parameters of the training dataset and the test cases of the contact benchmark.}
    \label{fig:contact_cases}
\end{figure}

Fig.~\ref{fig:contact_cases} shows the indenter parameters $a$ and $\gamma_a$ of the 50 training samples and the test cases of the contact benchmark. Table~\ref{tab:test_indenters} lists the parameters of the ten test indenters, numbered as in the figure. Test indenters 1--5 lie inside the training ranges, $a\in[160,\,320]$ mm and $\gamma_a\in[0.125,\,0.875]$, while test indenters 6--10 lie outside the range, making them extrapolation cases. The test indenters lie outside the training ranges by up to 29\% of the range width in $a$ and 9\% in $\gamma_a$.

\begin{table}[htbp]
    \centering
    \small\setlength{\tabcolsep}{3pt}\begin{tabular}{l|ccccc|ccccc}
        \hline
         & \multicolumn{5}{c|}{Inside the training ranges} & \multicolumn{5}{c}{Outside the training ranges} \\
        Test indenter & 1 & 2 & 3 & 4 & 5 & 6 & 7 & 8 & 9 & 10 \\
        \hline
        $a$ (mm) & 161.0 & 301.0 & 232.2 & 276.9 & 219.4 & 331.1 & 142.3 & 129.0 & 365.8 & 113.5 \\
        $\gamma_a$ & 0.308 & 0.486 & 0.577 & 0.126 & 0.746 & 0.945 & 0.939 & 0.088 & 0.073 & 0.360 \\
        \hline
    \end{tabular}
    \caption{Test indenters of the contact benchmark.}
    \label{tab:test_indenters}
\end{table}

\begin{figure}[htbp]
    \centering
    \includegraphics[width=\linewidth]{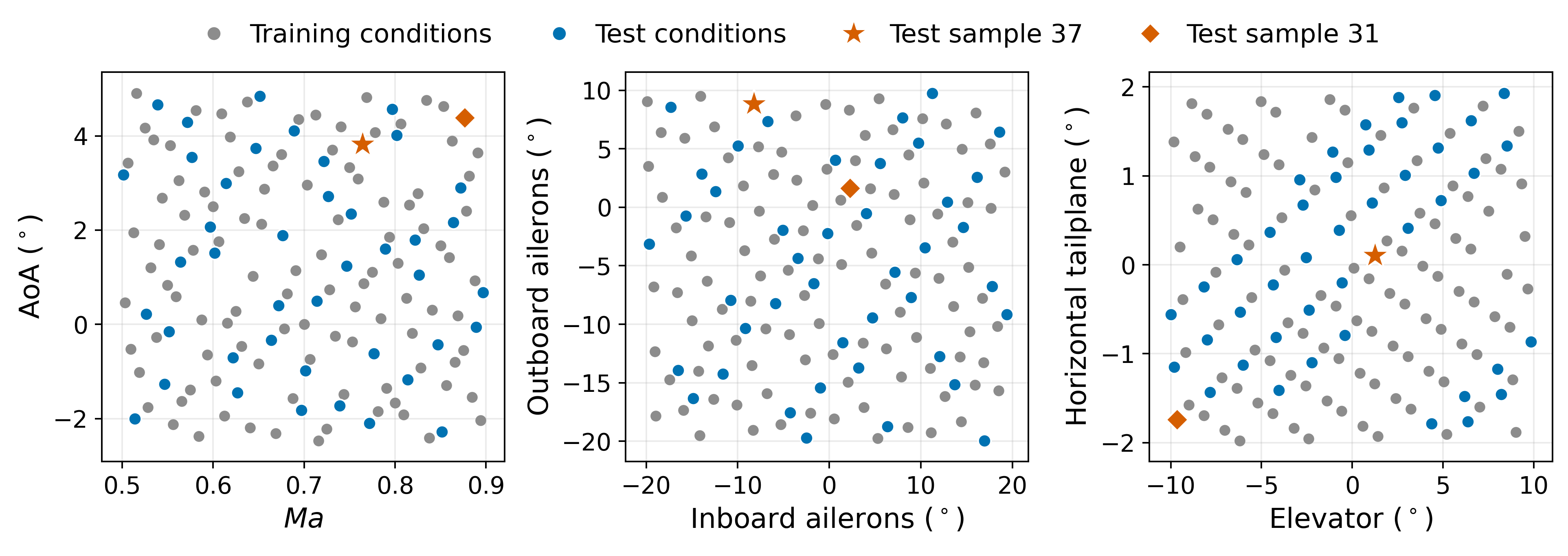}
    \caption{Training and test conditions of the NASA-CRM dataset.}
    \label{fig:crm_cases}
\end{figure}

Fig.~\ref{fig:crm_cases} shows the 105 training and 44 test conditions of the NASA-CRM dataset in three planes of the six input parameters, and Table~\ref{tab:crm_ranges} lists the range of each parameter. The test conditions are distributed over the same ranges as the training conditions. Test sample 37, shown in Figs.~\ref{fig:ex3_inference} and \ref{fig:NASA-CRM_inference}(a), and test sample 31, shown in Fig.~\ref{fig:NASA-CRM_inference}(b), are marked in Fig.~\ref{fig:crm_cases}, and their conditions are given in the last two columns of Table~\ref{tab:crm_ranges}.

\begin{table}[htbp]
    \centering
    \small\setlength{\tabcolsep}{4pt}\begin{tabular}{l|cc|cc}
        \hline
        Parameter & \hdtwo{Training}{conditions} & \hdtwo{Test}{conditions} & \hdtwo{Test}{sample 37} & \hdtwo{Test}{sample 31} \\
        \hline
        $Ma$ & $[0.503,\,0.894]$ & $[0.502,\,0.897]$ & 0.764 & 0.877 \\
        AoA ($^\circ$) & $[-2.47,\,4.91]$ & $[-2.28,\,4.85]$ & 3.83 & 4.38 \\
        Inboard ailerons ($^\circ$) & $[-19.88,\,19.18]$ & $[-19.65,\,19.42]$ & $-$8.22 & 2.27 \\
        Outboard ailerons ($^\circ$) & $[-19.76,\,9.52]$ & $[-19.95,\,9.76]$ & 8.85 & 1.65 \\
        Elevator ($^\circ$) & $[-9.83,\,9.67]$ & $[-9.98,\,9.83]$ & 1.25 & $-$9.65 \\
        Horizontal tailplane ($^\circ$) & $[-1.98,\,1.86]$ & $[-1.79,\,1.93]$ & 0.11 & $-$1.74 \\
        \hline
    \end{tabular}
    \caption{Parameter ranges of the NASA-CRM conditions and the two visualized test samples.}
    \label{tab:crm_ranges}
\end{table}

\clearpage
\bibliographystyle{elsarticle-num.bst} 
\bibliography{HI-MGN_refs}

@article{ostrowski2008solving,
  title={Solving inverse heat conduction problems using trained {POD-RBF} network inverse method},
  author={Ostrowski, Ziemowit and Bia{\l}ecki, Ryszard A and Kassab, Alain Jacques},
  journal={Inverse Problems in Science and Engineering},
  volume={16},
  number={1},
  pages={39--54},
  year={2008},
  publisher={Taylor \& Francis}
}

@article{guo2018reduced,
  title={Reduced order modeling for nonlinear structural analysis using {G}aussian process regression},
  author={Guo, Mengwu and Hesthaven, Jan S},
  journal={Computer Methods in Applied Mechanics and Engineering},
  volume={341},
  pages={807--826},
  year={2018},
  publisher={Elsevier}
}

@article{Hesthaven:2018a,
  title={Non-intrusive reduced order modeling of nonlinear problems using neural networks},
  author={Hesthaven, Jan S and Ubbiali, Stefano},
  journal={Journal of Computational Physics},
  volume={363},
  pages={55--78},
  year={2018},
  publisher={Elsevier}
}

@article{li2021nonintrusive,
  title={A nonintrusive parametrized reduced-order model for periodic flows based on extended proper orthogonal decomposition},
  author={Li, Teng and Deng, Shiyuan and Zhang, Kun and Wei, Haibo and Wang, Runlong and Fan, Jun and Xin, Jianqiang and Yao, Jianyao},
  journal={International Journal of Computational Methods},
  volume={18},
  number={9},
  pages={2150035},
  year={2021},
  publisher={World Scientific}
}

@article{kneifl2021nonintrusive,
  title={A nonintrusive nonlinear model reduction method for structural dynamical problems based on machine learning},
  author={Kneifl, Jonas and Grunert, Dennis and Fehr, Joerg},
  journal={International Journal for Numerical Methods in Engineering},
  volume={122},
  number={17},
  pages={4774--4786},
  year={2021},
  publisher={Wiley Online Library}
}

@article{mohan2018deep,
  title={A deep learning based approach to reduced order modeling for turbulent flow control using {LSTM} neural networks},
  author={Mohan, Arvind T and Gaitonde, Datta V},
  journal={arXiv preprint arXiv:1804.09269},
  year={2018}
}

@article{Lee:2024a,
  title={Parametric model order reduction by machine learning for fluid–structure interaction analysis},
  author={Lee, S. and Jang, K. and Lee, S. and Cho, H. and Shin, S. J.},
  journal={Engineering with Computers},
  volume={40},
  pages={45--60},
  year={2024},
  publisher={Springer}
}

@article{Lee:2021a,
  title={Parametric non-intrusive model order reduction for flow-fields using unsupervised machine learning},
  author={Lee, S. and Jang, K. and Cho, H. and Kim, H. and Shin, S. J.},
  journal={Computer Methods in Applied Mechanics and Engineering},
  volume={384},
  pages={113999},
  year={2021},
  publisher={Elsevier}
}

@article{Kadeethum:2022a,
  title={Non-intrusive reduced order modeling of natural convection in porous media using convolutional autoencoders: Comparison with linear subspace techniques},
  author={Kadeethum, Teeratorn and Ballarin, Francesco and Choi, Youngsoo and O’Malley, Daniel and Yoon, Hongkyu and Bouklas, Nikolaos},
  journal={Advances in Water Resources},
  volume={160},
  pages={104098},
  year={2022},
  publisher={Elsevier}
}

@article{Kadeethum:2022b,
  title={Reduced order modeling for flow and transport problems with {Barlow Twins} self-supervised learning},
  author={Kadeethum, Teeratorn and Ballarin, Francesco and O’Malley, Daniel and Choi, Youngsoo and Bouklas, Nikolaos and Yoon, Hongkyu},
  journal={Scientific Reports},
  volume={12},
  number={1},
  pages={20654},
  year={2022},
  publisher={Nature Publishing Group UK London}
}

@article{xu2020multi,
  title={Multi-level convolutional autoencoder networks for parametric prediction of spatio-temporal dynamics},
  author={Xu, Jiayang and Duraisamy, Karthik},
  journal={Computer Methods in Applied Mechanics and Engineering},
  volume={372},
  pages={113379},
  year={2020},
  publisher={Elsevier}
}

@article{lee2020model,
  title={Model reduction of dynamical systems on nonlinear manifolds using deep convolutional autoencoders},
  author={Lee, Kookjin and Carlberg, Kevin T},
  journal={Journal of Computational Physics},
  volume={404},
  pages={108973},
  year={2020},
  publisher={Elsevier}
}

@article{solera2023beta,
  title={$\beta$-Variational autoencoders and transformers for reduced-order modelling of fluid flows},
  author={Solera-Rico, Alberto and Sanmiguel Vila, Carlos and G{\'o}mez, M. A. and Wang, Yuning and Almashjary, Abdulrahman and Dawson, Scott T. M. and Vinuesa, Ricardo},
  journal={arXiv preprint arXiv:2304.03571},
  year={2023}
}

@article{kang2022physics,
  title={Physics-aware reduced-order modeling of transonic flow via $\beta$-variational autoencoder},
  author={Kang, Yu-Eop and Yang, Sunwoong and Yee, Kwanjung},
  journal={Physics of Fluids},
  volume={34},
  number={7},
  pages={076103},
  year={2022},
  publisher={AIP Publishing LLC}
}

@article{wang2021flow,
  title={Flow field prediction of supercritical airfoils via variational autoencoder based deep learning framework},
  author={Wang, Jing and He, Cheng and Li, Runze and Chen, Haixin and Zhai, Chen and Zhang, Miao},
  journal={Physics of Fluids},
  volume={33},
  number={8},
  pages={086108},
  year={2021},
  publisher={AIP Publishing LLC}
}

@article{Lee:2024b,
  title={Data-driven nonlinear parametric model order reduction framework using deep hierarchical variational autoencoder},
  author={Lee, S. and Lee, S. and Jang, K. and Cho, H. and Shin, S. J.},
  journal={Engineering with Computers},
  volume={40},
  pages={2385--2400},
  year={2024},
  publisher={Springer}
}

@article{lee2025physics,
  title={Physics-aware neural network-based parametric model-order reduction of the electromagnetic analysis for a coated component},
  author={Lee, SiHun and Kang, Seung-Hoon and Lee, SangMin and Shin, SangJoon},
  journal={Engineering with Computers},
  volume={41},
  pages={785--799},
  year={2025},
  publisher={Springer}
}

@inproceedings{qi2017pointnet,
  title={{PointNet}: Deep learning on point sets for {3D} classification and segmentation},
  author={Qi, Charles R and Su, Hao and Mo, Kaichun and Guibas, Leonidas J},
  booktitle={Proceedings of the IEEE conference on computer vision and pattern recognition},
  pages={652--660},
  year={2017}
}

@article{qi2017pointnet++,
  title={{PointNet++}: Deep hierarchical feature learning on point sets in a metric space},
  author={Qi, Charles Ruizhongtai and Yi, Li and Su, Hao and Guibas, Leonidas},
  journal={Advances in neural information processing systems},
  volume={30},
  year={2017}
}

@article{lu2021learning,
  title={Learning nonlinear operators via {DeepONet} based on the universal approximation theorem of operators},
  author={Lu, Lu and Jin, Pengzhan and Pang, Guofei and Zhang, Zhongqiang and Karniadakis, George Em},
  journal={Nature Machine Intelligence},
  volume={3},
  number={3},
  pages={218--229},
  year={2021},
  publisher={Nature Publishing Group UK London}
}

@article{li2023fourier,
  title={Fourier neural operator with learned deformations for {PDEs} on general geometries},
  author={Li, Zongyi and Huang, Daniel Zhengyu and Liu, Burigede and Anandkumar, Anima},
  journal={Journal of Machine Learning Research},
  volume={24},
  number={388},
  pages={1--26},
  year={2023}
}

@article{li2023geometry,
  title={Geometry-informed neural operator for large-scale {3D PDEs}},
  author={Li, Zongyi and Kovachki, Nikola and Choy, Chris and Li, Boyi and Kossaifi, Jean and Otta, Shourya and Nabian, Mohammad Amin and Stadler, Maximilian and Hundt, Christian and Azizzadenesheli, Kamyar and others},
  journal={Advances in neural information processing systems},
  volume={36},
  pages={35836--35854},
  year={2023}
}

@article{wu2024transolver,
  title={Transolver: A fast transformer solver for {PDEs} on general geometries},
  author={Wu, Haixu and Luo, Huakun and Wang, Haowen and Wang, Jianmin and Long, Mingsheng},
  journal={arXiv preprint arXiv:2402.02366},
  year={2024}
}

@article{luo2025transolver++,
  title={Transolver++: An accurate neural solver for {PDEs} on million-scale geometries},
  author={Luo, Huakun and Wu, Haixu and Zhou, Hang and Xing, Lanxiang and Di, Yichen and Wang, Jianmin and Long, Mingsheng},
  journal={arXiv preprint arXiv:2502.02414},
  year={2025}
}

@article{zhou2026transolver,
  title={Transolver-3: Scaling up transformer solvers to industrial-scale geometries},
  author={Zhou, Hang and Wu, Haixu and Shangguan, Haonan and Ma, Yuezhou and Weng, Huikun and Wang, Jianmin and Long, Mingsheng},
  journal={arXiv preprint arXiv:2602.04940},
  year={2026}
}

@article{alkin2025ab,
  title={{AB-UPT}: Scaling neural {CFD} surrogates for high-fidelity automotive aerodynamics simulations via anchored-branched universal physics transformers},
  author={Alkin, Benedikt and Bleeker, Maurits and Kurle, Richard and Kronlachner, Tobias and Sonnleitner, Reinhard and Dorfer, Matthias and Brandstetter, Johannes},
  journal={arXiv preprint arXiv:2502.09692},
  year={2025}
}

@article{pfaff2020learning,
  title={Learning mesh-based simulation with graph networks},
  author={Pfaff, Tobias and Fortunato, Meire and Sanchez-Gonzalez, Alvaro and Battaglia, Peter W},
  journal={arXiv preprint arXiv:2010.03409},
  year={2020}
}

@inproceedings{sanchez2020learning,
  title={Learning to simulate complex physics with graph networks},
  author={Sanchez-Gonzalez, Alvaro and Godwin, Jonathan and Pfaff, Tobias and Ying, Rex and Leskovec, Jure and Battaglia, Peter},
  booktitle={International conference on machine learning},
  pages={8459--8468},
  year={2020},
  organization={PMLR}
}

@inproceedings{dwivedi2022graph,
  title={Graph neural networks with learnable structural and positional representations},
  author={Dwivedi, Vijay Prakash and Luu, Anh Tuan and Laurent, Thomas and Bengio, Yoshua and Bresson, Xavier},
  booktitle={International Conference on Learning Representations},
  year={2022}
}

@inproceedings{yu2024learning,
  title={Learning flexible body collision dynamics with hierarchical contact mesh transformer},
  author={Yu, Youn-Yeol and Choi, Jeongwhan and Cho, Woojin and Lee, Kookjin and Kim, Nayong and Chang, Kiseok and Woo, Chang-Seung and Kim, Ilho and Lee, Seok-Woo and Yang, Joon-Young and Yoon, Sooyoung and Park, Noseong},
  booktitle={International Conference on Learning Representations},
  year={2024}
}

@article{fortunato2022multiscale,
  title={Multiscale {MeshGraphNets}},
  author={Fortunato, Meire and Pfaff, Tobias and Wirnsberger, Peter and Pritzel, Alexander and Battaglia, Peter},
  journal={arXiv preprint arXiv:2210.00612},
  year={2022}
}

@inproceedings{gao2019graph,
  title={Graph {U-Nets}},
  author={Gao, Hongyang and Ji, Shuiwang},
  booktitle={International conference on machine learning},
  pages={2083--2092},
  year={2019},
  organization={PMLR}
}

@article{lino2021simulating,
  title={Simulating continuum mechanics with multi-scale graph neural networks},
  author={Lino, Mario and Cantwell, Chris and Bharath, Anil A and Fotiadis, Stathi},
  journal={arXiv preprint arXiv:2106.04900},
  year={2021}
}

@article{lino2022multi,
  title={Multi-scale rotation-equivariant graph neural networks for unsteady {E}ulerian fluid dynamics},
  author={Lino, Mario and Fotiadis, Stathi and Bharath, Anil A and Cantwell, Chris D},
  journal={Physics of Fluids},
  volume={34},
  number={8},
  pages={087110},
  year={2022},
  publisher={AIP Publishing LLC}
}

@inproceedings{han2022predicting,
  title={Predicting physics in mesh-reduced space with temporal attention},
  author={Han, Xu and Gao, Han and Pfaff, Tobias and Wang, Jian-Xun and Liu, Li-Ping},
  booktitle={International Conference on Learning Representations},
  year={2022}
}

@article{lam2023learning,
  title={Learning skillful medium-range global weather forecasting},
  author={Lam, Remi and Sanchez-Gonzalez, Alvaro and Willson, Matthew and Wirnsberger, Peter and Fortunato, Meire and Alet, Ferran and Ravuri, Suman and Ewalds, Timo and Eaton-Rosen, Zach and Hu, Weihua and Merose, Alexander and Hoyer, Stephan and Holland, George and Vinyals, Oriol and Stott, Jacklynn and Pritzel, Alexander and Mohamed, Shakir and Battaglia, Peter},
  journal={Science},
  volume={382},
  number={6677},
  pages={1416--1421},
  year={2023},
  publisher={American Association for the Advancement of Science}
}

@inproceedings{cao2023efficient,
  title={Efficient learning of mesh-based physical simulation with bi-stride multi-scale graph neural network},
  author={Cao, Yadi and Chai, Menglei and Li, Minchen and Jiang, Chenfanfu},
  booktitle={International conference on machine learning},
  pages={3541--3558},
  year={2023},
  organization={PMLR}
}

@article{deshpande2024magnet,
  title={{MAgNET}: A graph {U-Net} architecture for mesh-based simulations},
  author={Deshpande, Saurabh and Bordas, St{\'e}phane P. A. and Lengiewicz, Jakub},
  journal={Engineering Applications of Artificial Intelligence},
  volume={133},
  pages={108055},
  year={2024},
  publisher={Elsevier}
}

@inproceedings{deng2025evomesh,
  title={{EvoMesh}: Adaptive physical simulation with hierarchical graph evolutions},
  author={Deng, Huayu and Zhu, Xiangming and Wang, Yunbo and Yang, Xiaokang},
  booktitle={International conference on machine learning},
  pages={13327--13346},
  year={2025},
  organization={PMLR}
}

@inproceedings{bazzi2026physics,
  title={Physics-informed coarsening for multigrid graph neural surrogates},
  author={Bazzi, Amir and Nemer, Ramy and Alves, Jos{\'e} and Hachem, Elie},
  booktitle={International conference on machine learning},
  pages={7179--7194},
  year={2026},
  organization={PMLR}
}

@inproceedings{lino2025learning,
  title={Learning distributions of complex fluid simulations with diffusion graph networks},
  author={Lino, Mario and Pfaff, Tobias and Thuerey, Nils},
  booktitle={International Conference on Learning Representations},
  year={2025}
}

@article{wurth2025diffusion,
  title={Diffusion-based hierarchical graph neural networks for simulating nonlinear solid mechanics},
  author={W{\"u}rth, Tobias and Freymuth, Niklas and Neumann, Gerhard and K{\"a}rger, Luise},
  journal={Advances in neural information processing systems},
  volume={38},
  year={2025}
}

@article{battaglia2018relational,
  title={Relational inductive biases, deep learning, and graph networks},
  author={Battaglia, Peter W and Hamrick, Jessica B and Bapst, Victor and Sanchez-Gonzalez, Alvaro and Zambaldi, Vinicius and Malinowski, Mateusz and Tacchetti, Andrea and Raposo, David and Santoro, Adam and Faulkner, Ryan and others},
  journal={arXiv preprint arXiv:1806.01261},
  year={2018}
}

@inproceedings{li2018deeper,
  title={Deeper insights into graph convolutional networks for semi-supervised learning},
  author={Li, Qimai and Han, Zhichao and Wu, Xiao-Ming},
  booktitle={Proceedings of the AAAI conference on artificial intelligence},
  volume={32},
  pages={3538--3545},
  year={2018}
}

@inproceedings{bekemeyer2025introduction,
  title={Introduction of applied aerodynamics surrogate modeling benchmark cases},
  author={Bekemeyer, Philipp and Hariharan, Nathan and Wissink, Andrew M and Cornelius, Jason},
  booktitle={AIAA SCITECH 2025 Forum},
  pages={0036},
  year={2025}
}

@article{alkin2024universal,
  title={Universal physics transformers: A framework for efficiently scaling neural operators},
  author={Alkin, Benedikt and F{\"u}rst, Andreas and Schmid, Simon and Gruber, Lukas and Holzleitner, Markus and Brandstetter, Johannes},
  journal={Advances in neural information processing systems},
  volume={37},
  year={2024}
}

@article{wen2025geometry,
  title={Geometry aware operator transformer as an efficient and accurate neural surrogate for {PDEs} on arbitrary domains},
  author={Wen, Shizheng and Kumbhat, Arsh and Lingsch, Levi and Mousavi, Sepehr and Zhao, Yizhou and Chandrashekar, Praveen and Mishra, Siddhartha},
  journal={Advances in neural information processing systems},
  volume={38},
  year={2025}
}
\end{document}